\documentclass[10pt,twocolumn,letterpaper]{article}

\usepackage{iccv}
\usepackage{times}
\usepackage{epsfig}
\usepackage{graphicx}
\usepackage{amsmath}
\usepackage{amssymb}
\usepackage{caption}

\usepackage{verbatim}
\usepackage{multirow}
\usepackage{booktabs}
\usepackage{bbm}
\usepackage{mathtools}
\usepackage{grffile}

\DeclarePairedDelimiter\floor{\lfloor}{\rfloor}

\newcommand*{\matminus}{%
  \leavevmode
  \hphantom{0}%
  \llap{%
    \settowidth{\dimen0 }{$0$}%
    \resizebox{1.1\dimen0 }{\height}{$-$}%
  }%
}

\newcommand\blfootnote[1]{%
  \begingroup
  \renewcommand\thefootnote{}\footnote{#1}%
  \addtocounter{footnote}{-1}%
  \endgroup
}
\newcommand*{\affmark}[1][*]{\textsuperscript{#1}}

\newcommand{\specialcell}[2][c]{%
  \begin{tabular}[#1]{@{}c@{}}#2\end{tabular}}

\usepackage[pagebackref=true,breaklinks=true,letterpaper=true,colorlinks,bookmarks=false]{hyperref}

\iccvfinalcopy % *** Uncomment this line for the final submission

\def\iccvPaperID{6958} % *** Enter the ICCV Paper ID here

\begin{document}

%%%%%%%%% TITLE
\title{Deep Edge-Aware Interactive Colorization against Color-Bleeding Effects}

\author{
  Eungyeup Kim\affmark[*1] \ Sanghyeon Lee\affmark[*1] \ Jeonghoon Park\affmark[*1] \ Somi Choi\affmark[1] \\ Choonghyun Seo\affmark[2] \ Jaegul Choo\affmark[1] \vspace{0.2cm}\\
  \affmark[1]KAIST, \ \affmark[2]NAVER WEBTOON Corp.\\
  \texttt{\footnotesize \affmark[1]\{eykim94, shlee6825, jeonghoon\texttt{\_}park, smchoi257, jchoo\}@kaist.ac.kr}, \texttt{\footnotesize \affmark[2]choonghyun.seo@webtoonscorp.com} \\
}

% \maketitle

\twocolumn[{
\maketitle
\vspace{-0.7cm}
\begin{center}
    \centering 
    \includegraphics[width=\linewidth]{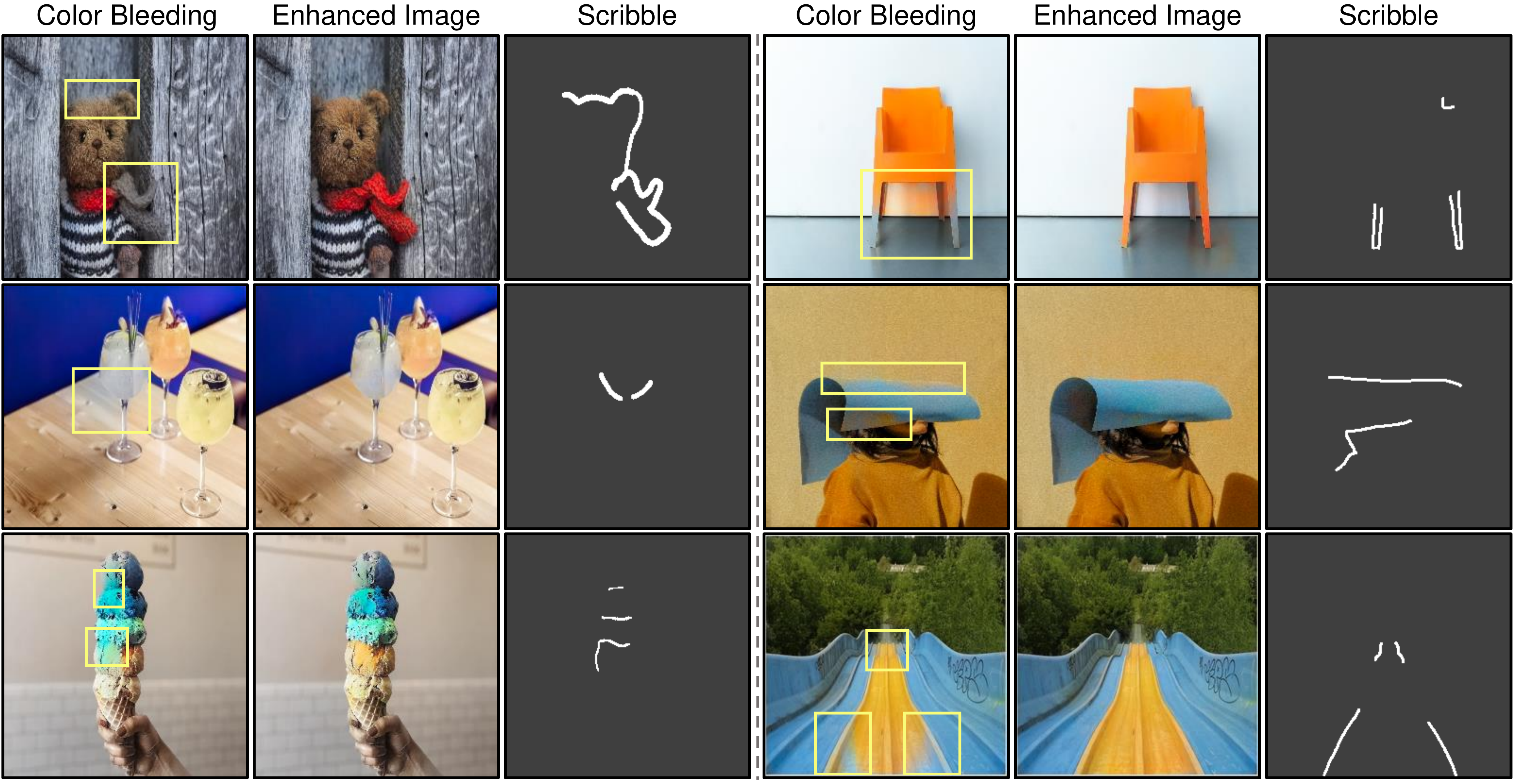}
    \vspace{-0.6cm}
    \captionof{figure}{
    Qualitative results of edge enhancement by using our proposed method. These images are collected from \url{https://unsplash.com}. The first and the fourth columns show samples containing color-bleeding artifacts. Yellow boxes represent color-bleeding regions. The second and the fifth columns have edge-enhanced samples by our proposed method. The scribbles utilized for edge enhancement are shown in the third and the sixth columns. Please see our \href{https://eungyeupkim.github.io/edge-enhancing-colorization/}{project webpage} for the demo video of our method.
    }
    \vspace{0.5cm}
\label{fig:qual_1}
\vspace{-0.4cm}
\end{center}
}]
\blfootnote{* indicates equal contribution\vspace{-0.8cm}}

\vspace{-0.4cm}
\begin{abstract}
\vspace{-0.2cm}
    Deep neural networks for automatic image colorization often suffer from the color-bleeding artifact, a problematic color spreading near the boundaries between adjacent objects. 
    Such color-bleeding artifacts debase the reality of generated outputs, limiting the applicability of colorization models in practice. 
    Although previous approaches have attempted to address  this problem in an automatic manner, they tend to work only in limited cases where a high contrast of gray-scale values are given in an input image.
    Alternatively, leveraging user interactions would be a promising approach for solving this color-breeding artifacts.
    In this paper, we propose a novel edge-enhancing network for the regions of interest via simple user scribbles indicating where to enhance.
    In addition, our method requires a minimal amount of effort from users for their satisfactory enhancement.
    Experimental results demonstrate that our interactive edge-enhancing approach effectively improves the color-bleeding artifacts compared to the existing baselines across various datasets.
\vspace{-0.4cm}
\end{abstract}

\section{Introduction}
\label{sec:01-introduction}

\begin{figure}[t!]
\begin{center}
  \includegraphics[width=\linewidth]{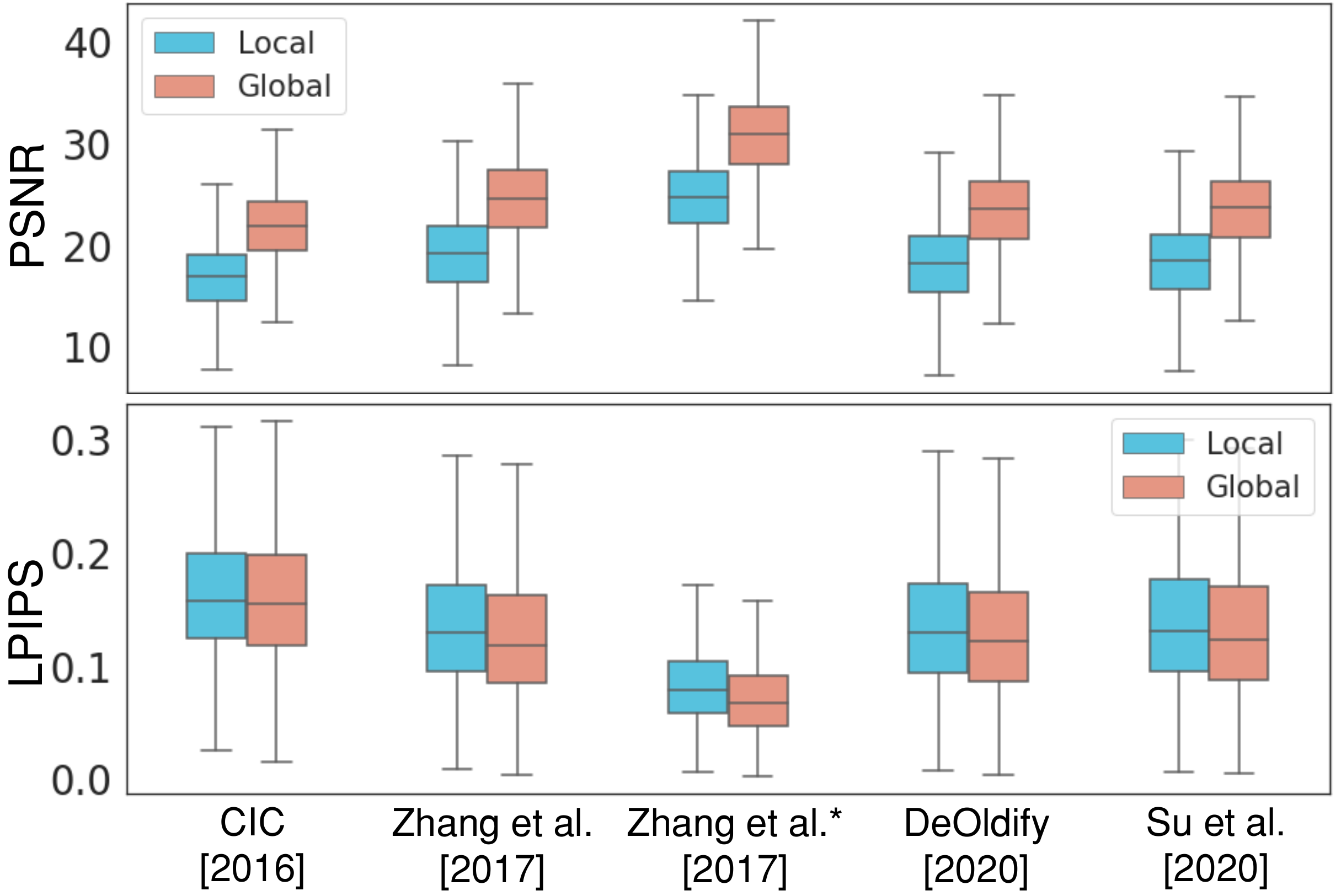}
  \end{center}
  \vspace{-0.4 cm}
  \caption{Comparison of local (around edges) and global PSNR and LPIPS scores of baseline models.
  Evaluations are conducted over ImageNet~\cite{imagenet_cvpr09} ctest10k dataset, and the edges are extracted from the ground-truth, using Canny edge extractor~\cite{canny1986cannyedge}.
  Throughout this paper, the symbol $*$ denotes the conditional model, which takes the color hints as additional input.
}
\label{fig:cb_boxplot}
\end{figure}

In recent years, deep image colorization methods~\cite{Iizuka2016letcolor, larsson2016learnrepC, zhang2016colorful, yoo2019memo, su2020insta, zhang2017real, zhang2018twostage, he2018deep, zhang2019deep, xiao2018interactive, lee2020refsketchC} have achieved a great performance on the generation of a realistic colorized image given a gray-scale or a sketch image.
However, these methods often contain the \textit{color-bleeding} artifact, a problematic color spreading across the adjacent objects.
As shown in Fig.~\ref{fig:qual_1}, the color-bleeding artifacts degrade the colorization quality particularly along the edges (red), compared to that in the whole image (blue).
For the quantitative analysis, we also compare the quality of the colorized outputs between the existing methods~\cite{zhang2016colorful, deoldify, zhang2017real, pixelated2020zhao, su2020insta} along the edges and the entire region in Fig.~\ref{fig:cb_boxplot}.
The result demonstrates that existing colorization methods, including Zhang \etal~\cite{zhang2017real} and Su \etal~\cite{su2020insta} which are widely used, suffer from the quality degradation particularly along the edges.
Therefore, we believe there is still room for further improvement in the colorization task by resolving such color-bleeding problem.

Some approaches have addressed this issue by applying a sharpening filter on an image to colorize~\cite{huang2005adaptiveedgeC, yin2019slide, Zhao2018PixellevelSG}, or leveraging additional tasks, such as semantic segmentation, to enhance the boundaries of the semantic objects~\cite{su2020insta, pixelated2020zhao}.
However, their improvements are limited to edges appearing with strong gray-scale contrast or along the objects corresponding to the predefined categories.
Color bleeding often occurs between the different objects that share similar gray-scale intensities and also along the edges that appear inside the objects, such as a zigzag pattern of a color pencil in Fig.~\ref{fig:qual_2}.
Therefore, tackling these bleeding edges at any desired locations still remains challenging, even in the recently proposed colorization methods.
Moreover, evaluation of the color bleeding regions can be highly subjective depending on the users, as the plausible boundaries of the multi-modal colorized objects can differ by point of view.

Therefore, we propose a novel interactive edge enhancement framework that takes a direct user interaction annotating a color-bleeding edge.
Unlike the previous approaches, our framework guarantees the reliable edge enhancement in any desired regions by utilizing user interactions.
In addition, our interactive approach only requires users the minimum efforts for edge enhancement.
We first apply a simple add-on edge-enhancing network, which takes both scribbles and an intermediate activation map of the colorization network as inputs.
This network encodes an edge-corrective representation for its input activation map, particularly in the regions annotated by the scribbles, and adds it into the original activation map by a residual connection.
Given this refined representation for the bleeding edges, the following layers of the colorization network can generate the edge-enhanced colorization output.

Experimental results demonstrate that our method has a remarkable performance over the baselines on diverse benchmark datasets, ImageNet~\cite{imagenet_cvpr09}, COCO-Stuff~\cite{caesar2018coco} and Place205~\cite{zhou2017places}.
Moreover, we introduce a new evaluation metric for measuring how reliably the colorization methods obey the color boundaries.
Also, we confirm that our approach takes the reasonable amount of time and efforts through the user-study, representing its potential in practical applications .
Furthermore, we explore the applicability of our approach in the task of sketch colorization as well, by validating our method on Yumi's Cells~\cite{yumicells} and Danbooru~\cite{danbooru2017} datasets.

\section{Related Work}
\label{sec:02-related_work}

\subsection{Unconditional and Conditional Colorization}
Deep learning-based colorization methods~\cite{Iizuka2016letcolor, zhang2016colorful, larsson2016learnrepC, Zhao2018PixellevelSG, Ho2020SemanticdrivenC, vitoria2020chromagan, yoo2019memo, su2020insta} have proposed fully automatic colorization approaches without any additional conditions.
These unconditional models predict the most plausible colors for the given input image, even without any laborious color annotations provided by a user.
By leveraging conditions given by the user, the recent colorization methods have accomplished multi-modal colorization.
One of the widely used conditions is a reference image~\cite{lee2020refsketchC, he2018deep, zhang2019deep}.
However, a reference image containing visually different contents from the gray-scale often induces implausible results. 
On the other hand, a color palette or a scribble hint given by a user directly designates the user's preference on both color and region~\cite{levin2004colorusingopt, zhang2017real, zhang2018twostage}.
However, as we observed in Fig.~\ref{fig:cb_boxplot}, both unconditional and conditional colorization often fail to preserve the color edge, resulting in generating color-bleeding artifacts along the boundary regions.

\subsection{Edge-Aware Colorization}
Classical approaches~\cite{huang2005adaptiveedgeC, yin2019slide} address the bleeding artifacts in an optimization problem.
Huang \etal~\cite{huang2005adaptiveedgeC} develop an edge detection algorithm for improving edges information during colorization.
Yin \etal~\cite{yin2019slide} propose a sharpening filter applied over a colorized image, alleviating the bleeding artifacts via optimization.
Similarly, Zhao \etal~\cite{Zhao2018PixellevelSG} propose a joint bilateral filter which considers the adjacent color values for sharpening the edges. 
However, as these approaches mainly rely on the edges of input image, they still fail on the boundaries between the objects that share the similar gray-scale intensities.
In contrast, our approach involves a direct interaction, which refines the boundary representation in any region regardless of its edge pixel values in the input image.

Recently, Su \etal~\cite{su2020insta} and Zhao \etal~\cite{pixelated2020zhao} leverage the semantic segmentation and object detection, respectively, when training a colorization model.
These tasks enforce the network to learn the semantic objects, which may help to recognize the boundary of such objects to colorize.
However, since both semantic segmentation and object detection recognize the objects defined by particular classes, the methods may still suffer from the color bleeding along the objects that are not classified by any categories.
For example, in Fig.~\ref{fig:qual_2}, the color-bleeding artifacts across the patterns inside the balloon would not be fully addressed by these methods, as such patterns are not classified into a certain class.
In our approach, we leverage the scribbles that are independent of object types, allowing more general edge enhancement against these methods.

\begin{figure}
\begin{center}
  \includegraphics[width=\linewidth]{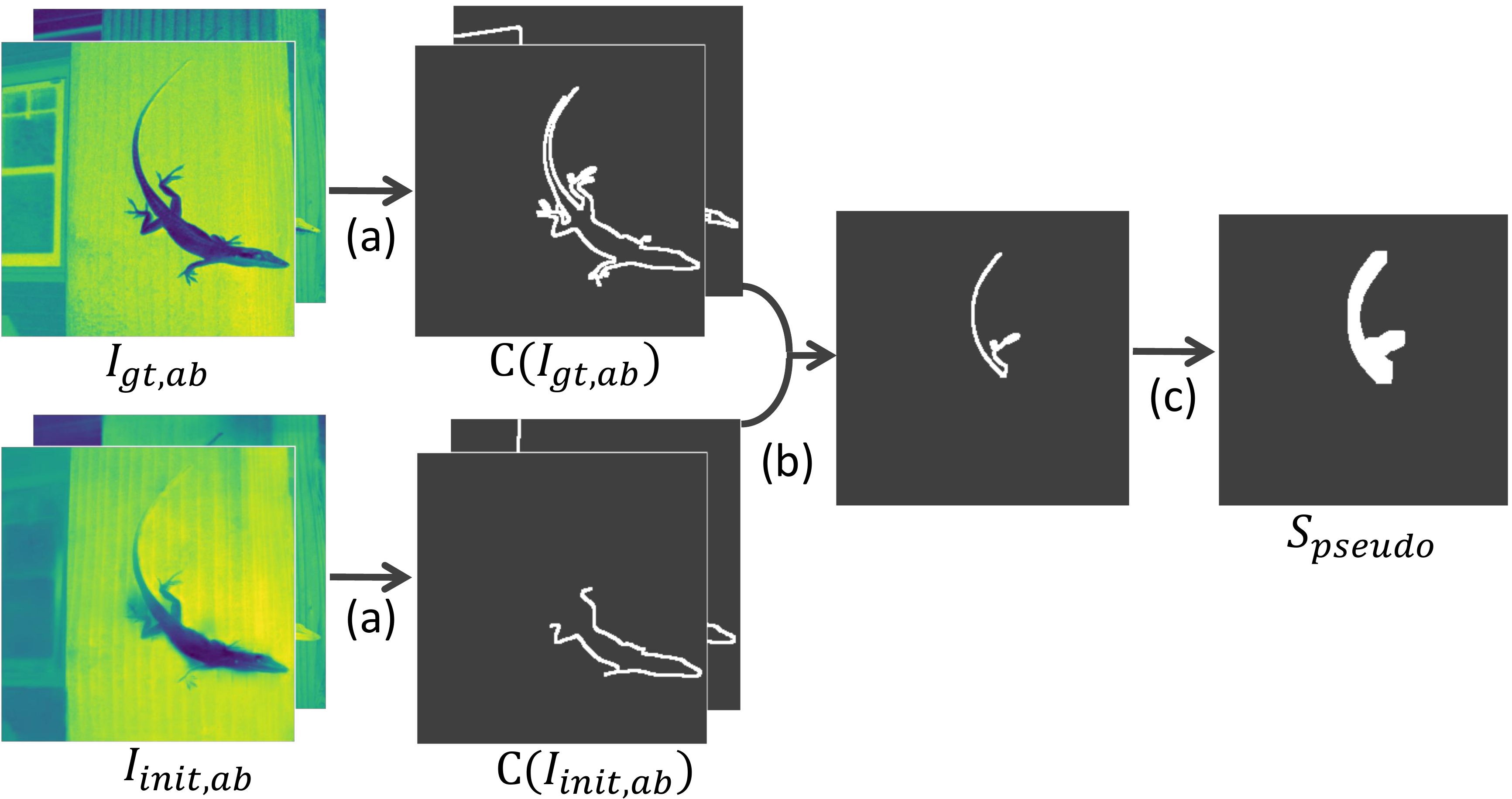}
  \end{center}
  \vspace{-0.4cm}
  \caption{An overview of the pseudo-scribble generation. Two channels of $I_{\text{gt},ab}$ and $I_{\text{init},ab}$ represent an $a$ and $b$ channel, respectively.}
\label{fig:pseudo-scribble}
\vspace{-0.3 cm}
\end{figure}

\begin{figure*}
\begin{center}
  \includegraphics[width=1\linewidth]{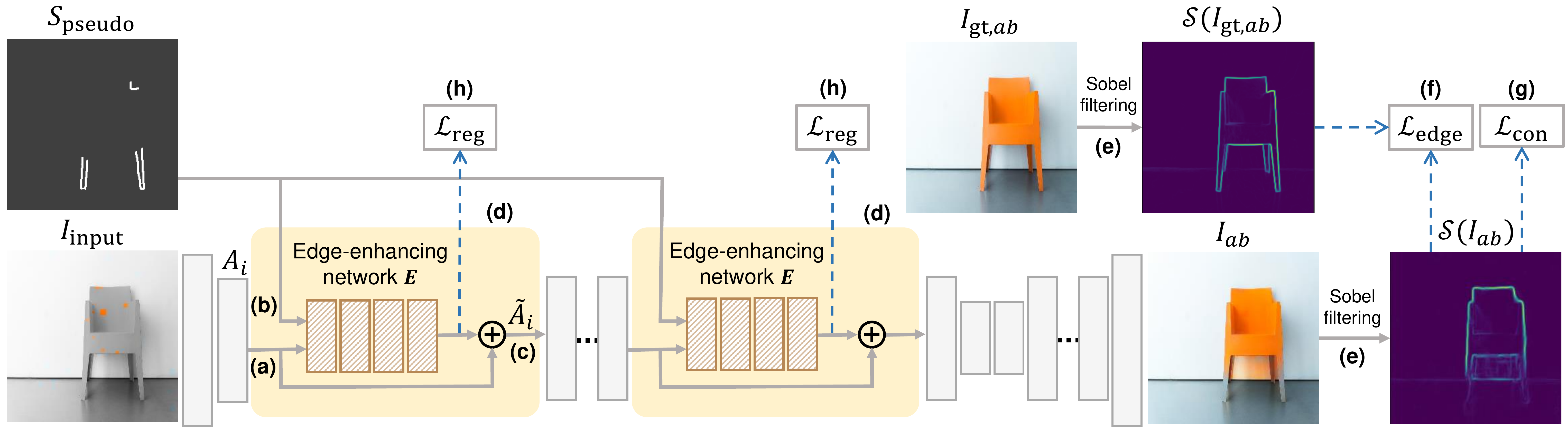}
  \end{center}
  \vspace{-0.6 cm}
  \caption{\textbf{An overview of our proposed method.} First, base colorization model (gray networks) colorize gray-scale image \(\textit{\textbf{I}}_{\text{input}}\). 
  After, multiple add-on edge-enhancing network \textit{\textbf{E}} (yellow boxes) take the user-driven scribble \(\textit{\textbf{S}}_{\text{pseudo}}\) and refine the corresponding activation maps from the base model.
  We apply \(\mathcal{\textit{\textbf{L}}}_{\text{edge}}\) between the edges of \(\textit{\textbf{I}}_{ab}\) and \(\textit{\textbf{I}}_{\text{gt}, ab}\). }
\label{fig:method_overview}
\vspace{-0.3cm}
\end{figure*}

\section{Proposed Method}
\label{sec:03-proposed_method}
\subsection{Overall Workflow}
This section provides a detailed description of our proposed method, as described in Fig.~\ref{fig:method_overview}.
As an interactive approach, we design an edge-enhancing network $E$ to take scribbles, which annotates the color-bleeding edges, as additional inputs.
These scribbles, which we term pseudo-scribbles, are automatically generated to approximate the real-world user hints (Section~\ref{Pseudo-Scribble}).
The network $E$ also takes the intermediate activation maps of the colorization network, and refines them along the edges annotated by the pseudo-scribbles.
Afterward, these refined representations pass through the following layers to obtain an edge-enhanced colorized output in the end (Section~\ref{edge_refinement_network}).
To train our model, we introduce an edge-enhancing loss, which enforces the model to recover the clear edges close to those of the ground-truth image.
Also, we propose both feature-regularization loss and consistency loss to prevent undesirable color distortion which debases the overall quality of edge-enhanced outputs (Section~\ref{sec:objective_functions}).

\subsection{Pseudo-Scribble}
\label{Pseudo-Scribble}
Our approach requires a user-driven hint to be trained in an interactive manner, but collecting real-world user annotations needs a lot of time and human resources, which is prohibitive.
Instead, we automatically generate the pseudo-scribbles $S_{\text{pseudo}}$, which emulate the real user scribbles, for each training image.
The overall procedure of generating the $S_{\text{pseudo}}$ is presented in Fig.~\ref{fig:pseudo-scribble}. 
First, we obtain the color-bleeding outputs $I_\text{init}$ from a pre-trained colorization model.
Afterwards, we apply the Canny edge detector~\cite{canny1986cannyedge} $C(\cdot)$, a widely used edge detecting algorithm, onto the $\textit{ab}$ color channels of a ground-truth image $I_{\text{gt}}$ and a $I_\text{init}$, respectively (Fig.~\ref{fig:pseudo-scribble} \textbf{(a)}).
Then, we can obtain the binary maps $C(I_{\text{gt}, ab})$ and $C(I_{\text{init}, ab})$ which represent the edges. 

By selecting one of the edges that appears in $C(I_{\text{gt},ab})$, but not in $C(I_{\text{init},ab})$, we can have a single edge where the baseline fails to preserve boundary as clearly as $I_{\text{gt}}$ does (Fig.~\ref{fig:pseudo-scribble} \textbf{(b)}).
Note that while a gecko's tail is shown to be selected in \textbf{(b)}, other scribbles can be chosen in the training as well, such as its paw. 
Afterward, to better approximate a real user's hint, the $S_{\text{pseudo}}$ are formed to be thick and coarse enough to $1$) be easy to draw and $2$) contain the bleeding boundary.
To this end, we apply a width transformation $w(\cdot)$ that randomly modifies the width of the selected edge between 1 and 11 pixels (Fig.~\ref{fig:pseudo-scribble} \textbf{(c)}).

\subsection{Edge-Enhancing Network}
\label{edge_refinement_network}
We apply an edge-enhancing network $E$ to refine the intermediate representations of a colorization network, correcting the erroneously spread colors across the boundaries.
This network encodes the corrective representations from both scribbles and the intermediate features as inputs and adds them to the original features with the residual connection.
Suppose that we want to modify an activation map $A_i$, where $A=(A_1, A_2, ...A_l)$ is the set of intermediate activation maps from $l$ different encoder layers of the colorization model (Fig.~\ref{fig:method_overview} (a)).
To obtain the scribbles, we generate a $S_{\text{pseudo}}$ by the procedure described in Section~\ref{Pseudo-Scribble} and downscale it to match the spatial resolution of the activation map $A_i$ from the $i$-th layer (Fig.~\ref{fig:method_overview} (b)).
Then, $S_{\text{pseudo}}$ and $A_i$ are concatenated to provide an input for edge-enhancing network $E$.
Given the concatenated tensor, we can obtain an activation map representing the correction of $A_i$ to alleviate color-bleeding artifacts.
Therefore, by applying a residual connection with $A_i$, (Fig.~\ref{fig:method_overview} (c)) refined activation map $\Tilde{A}_i$ is calculated as
\begin{equation}
    \Tilde{A}_i = A_i + E\left(\left[S_{\text{pseudo}}, A_i\right]\right),
\end{equation}
where $E$ is the proposed edge-enhancing network and $[\cdot , \cdot]$ a concatenation.
 
We apply the edge-enhancing networks to the encoder because the edge-enhancing performance is empirically better when applying our network \textit{E} to the encoder than the decoder.
Detailed comparisons on the qualitative results of the network \textit{E} applied in the encoder and decoder layer are provided in Section~\ref{sec:layers}.
To encourage edge refinement in both low- and high-level representations, we apply multiple edge-enhancing networks in both shallow and deep layers of the encoder (Fig.~\ref{fig:method_overview} (d)).

\begin{table*}
\begin{center}
\begin{tabular}{@{}lclllclllclll@{}}
\toprule
\multicolumn{1}{c}{\multirow{2}{*}{Kernel Size}} & \multicolumn{1}{c}{\multirow{2}{*}{Methods}} \phantom{abc} &
\multicolumn{2}{c}{ImageNet ctest~\cite{imagenet_cvpr09}} & \phantom{ab} &
\multicolumn{2}{c}{COCO-Stuff~\cite{caesar2018coco}} & \phantom{ab} &
\multicolumn{2}{c}{Place205~\cite{zhou2017places}} & \\
\cmidrule{3-4} \cmidrule{6-7} \cmidrule{9-10}
&& \multicolumn{1}{c}{LPIPS$\downarrow$} & \multicolumn{1}{c}{PSNR$\uparrow$} && \multicolumn{1}{c}{LPIPS$\downarrow$} & \multicolumn{1}{c}{PSNR$\uparrow$} && \multicolumn{1}{c}{LPIPS$\downarrow$} & \multicolumn{1}{c}{PSNR$\uparrow$} \\
\midrule
\multicolumn{1}{c}{\multirow{8}{*}{K=7}}
& CIC ~\cite{zhang2016colorful} & 0.248 & 13.281 && 0.247 & 13.368 && 0.254 & 13.577\\
& DeOldify ~\cite{deoldify} & 0.250 & 13.234 && 0.251 & 13.059 && 0.227 & 14.258\\
& Zhang \etal~\cite{zhang2017real}& 0.246 & 13.248 && 0.206 & 14.755 && 0.219 & 14.815\\
& +Ours  & \textbf{0.217}& \textbf{13.919} && \textbf{0.192} & \textbf{15.037} && \textbf{0.211} & \textbf{15.104}\\\cline{2-11}
& Zhang \etal~\cite{zhang2017real}$^*$& 0.208 & 14.966 && 0.158 & 17.456 && 0.171 & 17.530\\
& +Ours$^*$ & \textbf{0.177} & \textbf{16.041} && \textbf{0.143} & \textbf{17.953} && \textbf{0.161} & \textbf{17.906}\\ \cline{2-11}
& Su \etal~\cite{su2020insta}$^*$& 0.185 & 16.393 && 0.187 & 15.971 && 0.194 & 17.032 \\
& +Ours$^*$ & \textbf{0.177} & \textbf{16.507} && \textbf{0.176} & \textbf{16.188} && \textbf{0.187} & \textbf{17.098}\\
\midrule
\multicolumn{1}{c}{\multirow{8}{*}{K=Full}}
& CIC ~\cite{zhang2016colorful}& 0.172 & 21.001 && 0.164 & 21.456 && 0.153 & 21.873\\
& DeOldify ~\cite{deoldify} & 0.159 & 21.433 && 0.149 & 21.985 && 0.156 & 21.933\\
& Zhang \etal~ \cite{zhang2017real} & 0.148 & 21.981 && 0.135 & 22.729 && 0.138 & \textbf{22.846}\\
& +Ours & \textbf{0.147} & \textbf{22.026} && \textbf{0.134} & 22.729 && 0.138 & 22.845\\\cline{2-11}
& Zhang \etal~\cite{zhang2017real}$^*$& 0.086 & 27.202 && 0.080 & 27.681 && 0.087 & 27.697\\
& +Ours $^*$& \textbf{0.085} & \textbf{27.559} && \textbf{0.078} & \textbf{27.955} && 0.087 &
\textbf{27.935}\\ \cline{2-11}
& Su \etal~\cite{su2020insta}$^*$& 0.091 & 26.211 && 0.089 & 26.050 && 0.090 & 27.414\\
& +Ours$^*$ & 0.091 & \textbf{26.291} && \textbf{0.088} & \textbf{26.233} && \textbf{0.089} & \textbf{27.486}\\
\bottomrule
\end{tabular}
\end{center}
\vspace{-0.5cm}
\caption{Quantitative comparison with the baselines on 1,000 images in the ImageNet~\cite{imagenet_cvpr09}, COCO-Stuff~\cite{caesar2018coco} and Place205~\cite{zhou2017places} validation set. Quantitative results in the local region show that our method effectively enhances the images.} 
\label{tab:uncon_comp}
\vspace{-0.5cm}
\end{table*}

\subsection{Objective Functions}
\label{sec:objective_functions}
\noindent \textbf{Edge-Enhancing Loss.}
Inspired by the gradient difference loss (GDL)~\cite{Mathieu2016gdl} which sharpens the video prediction, we propose an edge-enhancing loss $\mathcal{L}_\text{edge}$ for enhancing the edges in a target region.
This loss  $\mathcal{L}_\text{edge}$ enforces an edge-enhancing network $E$ to generate the refined activation maps that enhance the edges of $I_{ab}$ to be close to those of $I_{\text{gt},ab}$ (Fig.~\ref{fig:method_overview} (f)).
To obtain the edge map, we utilize the Sobel filter, a differentiable edge extracting filter, onto the \textit{CIE ab} channels of images, obtaining both horizontal and vertical derivative approximations of color intensities (Fig.~\ref{fig:method_overview} (e)).

The resulting color gradient is formally written as
\addtolength{\jot}{2pt}
\begin{equation}
\label{eq1}
  \begin{split}
  &\mathcal{S}(I)= \sqrt{(G_{x} * I)^2 + (G_{y} * I)^2},\\
    &G_x= \begin{pmatrix}
        1 & 0 & \matminus1 \\
        2 & 0 & \matminus2 \\
        1 & 0 & \matminus1
        \end{pmatrix},\hspace{0.1cm} 
    G_y= \begin{pmatrix}
        1 & 2 & 1 \\
        0 & 0 & 0 \\
        \matminus1 & \matminus2 & \matminus1
        \end{pmatrix},
  \end{split} 
\end{equation}
where $G_x$ and $G_y$ are horizontal and vertical Sobel filters which convolve with the given image, and $\mathcal S$ returns the gradient magnitude of them.
Our proposed edge-enhancing loss can be written as 
\begin{equation}
    \begin{split}
    &\mathcal{L}_{\text{edge}}= \mathbb{E}_{x, y \in \mathbb{P}} \left[||S(x, y) - S_{\text{gt}}(x,y)||^2_2 \right],\\
    &S=\mathcal{S}(I_{ab}), \hspace{0.1cm} S_{\text{gt}}=\mathcal{S}(I_{\text{gt}, ab}),
    \end{split}
\end{equation} 
where $\mathbb{P}$ denotes a set of coordinates $\left(x, y\right)$ whose values are non-zero in a binary mask $M$.
$M$ only activates a set of pixels within certain distance from the target edge, \textit{i.e.}, $S_{\text{pseudo}}$.

\noindent \textbf{Feature-Regularization Loss.}
We wish for our proposed method to improve the edges while maintaining the original performance of the colorization network.
Therefore, we introduce a feature-regularization loss $\mathcal{L}_{\text{reg}}$ (Fig.~\ref{fig:method_overview} (h)) to the output of the edge-enhancing network $E$.
This encourages our network to learn optimal edge enhancement while avoiding the excessive perturbations in the network activation maps.
This loss is formulated as
\begin{equation}
    \mathcal{L}_{\text{reg}_i}= \lVert E\left(\left[S_{\text{pseudo}}, A_i\right]\right)\rVert^2_2,
\end{equation}
where $i$ denotes an index of a layer to be revised and $[\cdot , \cdot]$ a concatenation.

\noindent \textbf{Consistency Loss.}
While our proposed network enhances the gradient of colors around the given edges, it can unintentionally induce color distortions in the undesirable regions, \textit{i.e.}, outside of the target edges.
Therefore, we design an additional constraint, named consistency loss $\mathcal{L}_\text{con}$ (Fig.~\ref{fig:method_overview} (g)), to prevent these unnecessary changes.
This loss further optimizes our network $E$ to learn the refinements only in the desired regions.
$\mathcal{L}_\text{con}$ penalizes the unfavorable changes of our enhanced output $I_{ab}$ from the initial colorized output $I_{\text{init}, ab}$, only in the regions where we wish for the colors to remain.
As the binary mask $M$ mentioned above indicates the regions for the colors to be changed by edge enhancement, we apply this loss on the pixels whose values of $M$ are zero.
This is enabled by multiplying $1-M$ on each channel.
This loss can be formulated as 
\begin{equation}
    \begin{split}
        &\mathcal{L}_{\text{con}}= \mathbb{E}_{x, y \not\in \mathbb{P}} \left[\lVert S(x, y) - S_{\text{init}}(x,y)\rVert_2^2 \right],\\
        &S=\mathcal{S}(I_{ab}), \hspace{0.1cm} S_{\text{init}}=\mathcal{S}(I_{\text{init}, ab}),
    \end{split}
\end{equation}
where $\mathbb{P}$ denotes a set of coordinates $\left(x, y\right)$ whose values are non-zero in a binary mask $M$. 
\begin{figure*}
\begin{center}
  \includegraphics[width=\linewidth]{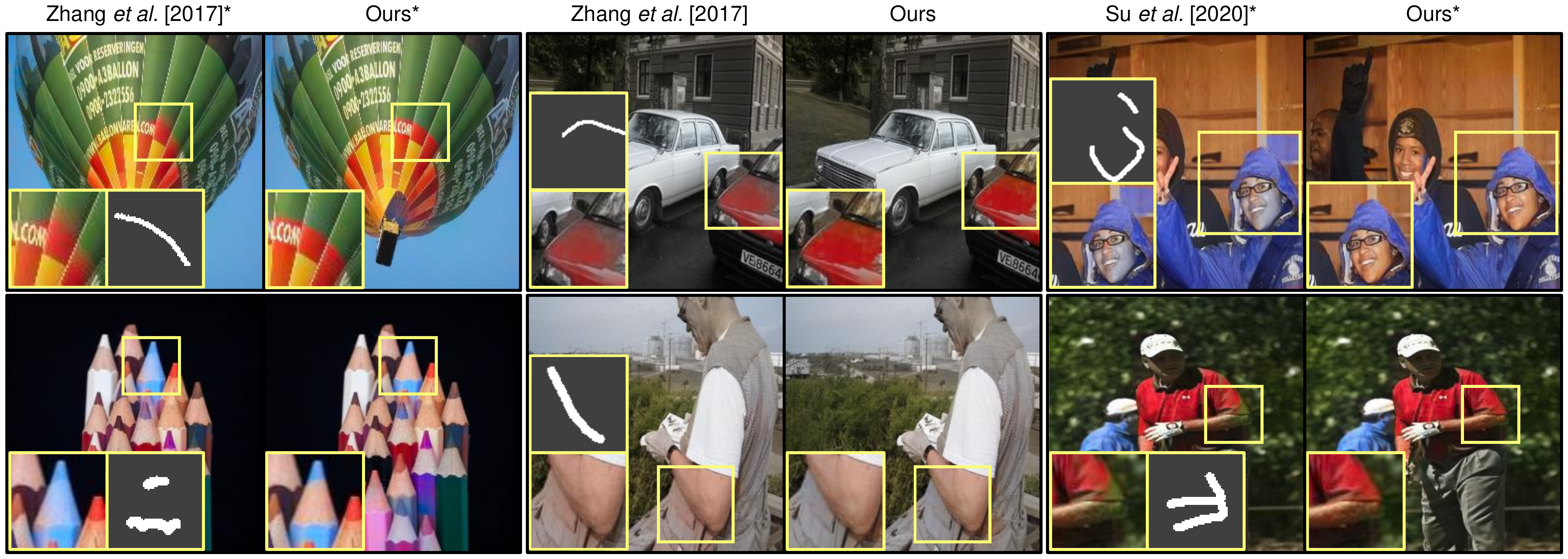}
  \end{center}
  \vspace{-0.6 cm}
  \caption{Qualitative examples of edge enhancement in gray-scale colorization. For each box, the left contains an original colorized image with artifacts, its magnified view, and given user scribble.}
\label{fig:qual_2}
\vspace{-0.4cm}
\end{figure*}

In summary, the overall objective function for training the edge-enhancing network is defined as 
\begin{equation} 
    \mathcal{L}_{\text{total}} = \lambda_{\text{edge}}\mathcal{L}_{\text{edge}} + \sum_{i=1}^l \lambda_{\text{reg}_i} \mathcal{L}_{\text{reg}_i} + \lambda_{\text{con}} \mathcal{L}_{\text{con}}, 
\end{equation}
where $\lambda_{\text{edge}}$, $\lambda_{\text{reg}}$ and $\lambda_{\text{con}}$ are hyperparameters, and $l$ is the number of layers with the edge-enhancing network.

\subsection{Implementation Details}
\label{implementation_details}
In our experiments, we use the colorization networks introduced in Zhang \etal~\cite{zhang2017real} and Su \etal~\cite{su2020insta} as our backbone colorization models.
Zhang \etal first proposes an interactive colorization approach that takes color hints, achieving state-of-the-art performance over the existing conditional methods.
Su \etal achieves superior performance in an unconditional setting by leveraging an object detection module for an instance-level colorization.
We empirically confirm that applying our framework to the baselines taking explicit color hints results in better optimized edge-enhancing networks, compared to training on unconditional ones.
Therefore, similar to Zhang \etal, we re-implement Su \etal to take local color hints as additional inputs, which is not available in the original paper. 
Further details are provided in Section~\ref{sec:implementation_details}.

\begin{table}
\vspace{+0.2cm}
\begin{center}
\begin{tabular}{@{}lclllclll@{}}
\toprule
\multicolumn{1}{c}{\multirow{2}{*}{Method}} &
\multicolumn{3}{c}{Cluster Discrepancy Ratio$\uparrow$} \\
\cmidrule{2-4}
& \specialcell{ImageNet\\ctest~\cite{imagenet_cvpr09}} & \specialcell{COCO-\\Stuff~\cite{caesar2018coco}} & \specialcell{Place205\\~\cite{zhou2017places}} \\
\midrule
CIC ~\cite{zhang2016colorful} & 0.383 & \multicolumn{1}{c}{0.401} & \multicolumn{1}{c}{0.381} \\
DeOldify ~\cite{deoldify} & 0.437 & \multicolumn{1}{c}{0.445} & \multicolumn{1}{c}{0.441}\\\cline{2-4}
Zhang \etal~\cite{zhang2017real} & 0.385 & \multicolumn{1}{c}{0.391} & \multicolumn{1}{c}{0.377}\\
+ Ours& \textbf{0.502} & \multicolumn{1}{c}{\textbf{0.521}} & \multicolumn{1}{c}{\textbf{0.473}}  \\
\midrule
Zhang \etal~\cite{zhang2017real}$^*$ & 0.418 & \multicolumn{1}{c}{0.421} & \multicolumn{1}{c}{0.402} \\
+ Ours $^*$& \textbf{0.543} & \multicolumn{1}{c}{\textbf{0.547}} & \multicolumn{1}{c}{\textbf{0.508}}  \\\cline{2-4}
Su \etal~\cite{su2020insta}$^*$ & 0.336 & \multicolumn{1}{c}{0.325} & \multicolumn{1}{c}{0.336} \\
+ Ours $^*$& \textbf{0.394} & \multicolumn{1}{c}{\textbf{0.398}} & \multicolumn{1}{c}{\textbf{0.371}}  \\
\bottomrule
\end{tabular}
\end{center}
\vspace{-0.5cm}
\caption{Quantitative results using cluster discrepancy ratio measured within the kernel size of 7 along the edges.
The score ranges from $0$ to $1$.}
\label{tab:cd_ratio}
\vspace{-0.4cm}
\end{table}

\section{Experiments}
\label{sec:04-experiments}
\noindent \textbf{Baselines.}
We compare our proposed model with various colorization methods, including both unconditional and conditional ones.
Unconditional baselines include CIC~\cite{zhang2016colorful}, DeOldify~\cite{deoldify}, and Zhang \etal without color hints.
For conditional baselines, we utilize Zhang \etal and Su \etal with local hints, as mentioned in Section~\ref{implementation_details}.

\noindent \textbf{Dataset.}
The experiments are conducted with dataset including \textit{ImageNet}~\cite{imagenet_cvpr09}, \textit{COCO-Stuff}~\cite{caesar2018coco} and \textit{Place205}~\cite{zhou2017places}, which are generally used in colorization tasks.

\noindent \textbf{Evaluation Measure.}
For evaluation, we assess the performance of our proposed method and other prior models by utilizing two measures, peak signal-to-ratio (PSNR) and learned perceptual image patch similarity (LPIPS)~\cite{zhang2018perceptual}.
In addition, we newly propose a metric named \textit{cluster discrepancy ratio} (CDR), which is designed to measure the degree of color-bleeding effects.

\begin{figure*}
\begin{center}
  \includegraphics[width=\linewidth]{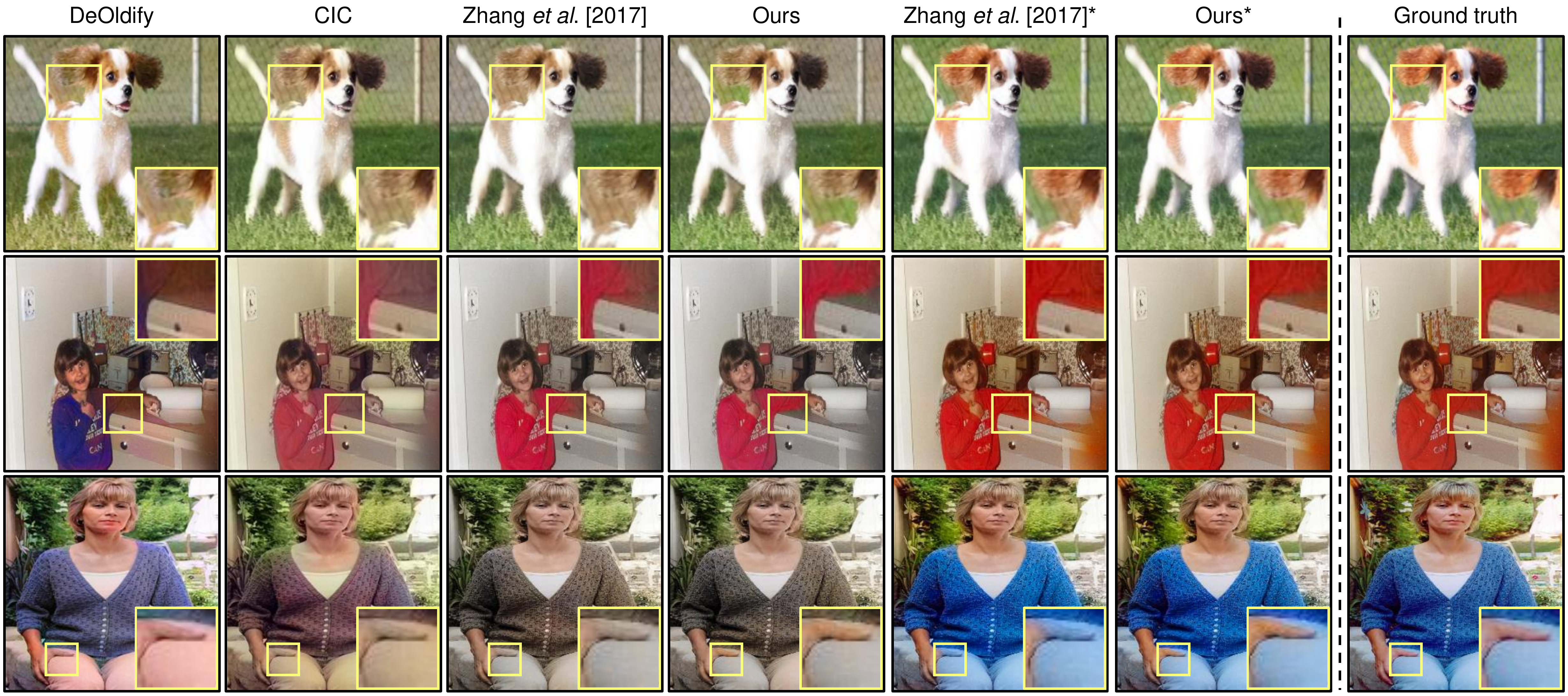}
  \end{center}
  \vspace{-0.6 cm}
  \caption{Qualitative comparisons between our model and baseline models. These images include color-bleeding in the same region for all baselines. Qualitative results show that our model successfully refinement this bleeding.}
\label{fig:qual_3}
\vspace{-0.4cm}
\end{figure*}

\subsection{Cluster Discrepancy Ratio}
\label{sec:ratio}
Although PSNR and LPIPS are generally used for evaluating the colorization performance in a rich literature~\cite{zhang2016colorful, zhang2017real, su2020insta, yoo2019memo, larsson2016learnrepC, zhang2019deep, lee2020refsketchC, Zhao2018PixellevelSG}, they are essentially based on the color difference between a generated image and a ground-truth one.
Therefore, a colorized image that contains well-preserved edges but different colors from the ground-truth may be underrated by these two metrics.
Note that this image would appear even more realistic compared to a bleeding image with similar colors.
This specific failure case is described with an example and its scores in Section~\ref{sec:discrepancy_ratio}.

To compensate for this concern, we propose a novel evaluation metric that measures the discrepancy of color clusters grouped by the super-pixels defined by a simple linear iterative clustering method~\cite{achanta2012slic}. 
The super-pixels have their cluster assignments $C$ based on color similarity. 
Inspired by this, we can perform binary classification on whether two adjacent pixels with different cluster assignments in the ground-truth still have different color values in the colorized outputs, especially along the edges.
Specifically, for the pixel $x_{\text{gt}}^{i}$ in a set of coordinates along the boundary $E$ of the ground-truth $I_{\text{gt},ab}$, we identify whether the adjacent pixel $x_{\text{gt}}^{j}$ within kernel size have different cluster assignments from that of $x_{\text{gt}}^{i}$.
For those who have different cluster assignments from $C_{x_{\text{gt}}^{i}}$, the cluster index of $x_{\text{gt}}^{i}$, we define a set $\Omega(i)$ that consists of their coordinates.
Then, in the generated outputs $I_{ab}$, we count the number of adjacent pixels $x^{j}$ that $1)$ belong to $\Omega(i)$ and $2)$ have the same cluster assignment as $C_{x^{i}}$.
The number indicates how many pixels are from different clusters in the $I_{\text{gt},ab}$, but share the same colors in the $I_{ab}$, which corresponds to the color-bleeding artifacts.
Therefore, the super-pixel-based cluster discrepancy ratio can be written as 
\begin{equation}
    \begin{split}
    &R\left(I_0, I_{\text{gt}}\right)=\frac{1}{\lvert E \rvert}\sum_{i\in E}\left(1 - \frac{1}{\lvert \Omega(i) \rvert}\sum_{k\in \Omega(i)} \mathbbm{1}_{C_{x^{i}} = C_{x^{i + k}}}\right),\\
    &\Omega(i)=\left\{j:C_{x_{\text{gt}}^{i}} \neq C_{x_{\text{gt}}^{i + j}}, j\in S \right\},
    \end{split}
\end{equation}
where $E$ denotes a set of coordinates for the pixels of edges, $C_{x_{\text{gt}}}$ and $C_{x}$ a cluster assignment given to the super-pixels of the $I_{\text{gt},ab}$ and $I_{ab}$.
All possible shifts $S$ within the kernel size $K$ are described as 
\begin{equation}
    S \coloneqq \left[ - \floor*{\frac{K}{2}},...,\floor*{\frac{K}{2}} \right] \times  \left[ - \floor*{\frac{K}{2}},...,\floor*{\frac{K}{2}} \right].
\end{equation}

\vspace{0.2cm}
\subsection{Qualitative Results}
\label{sec:qualitative_results}
In Fig.~\ref{fig:qual_1}, we visualize the images having color-bleeding artifacts from the conditional colorization model of Zhang \etal~\cite{zhang2017real} and their enhanced results using our proposed method.
This demonstrates that our approach robustly corrects the bleeding boundaries even when roughly drawn scribbles are given.
In addition, multiple edges can be enhanced in a single feed-forward when their corresponding scribbles are given at once.
Fig.~\ref{fig:qual_2} provides additional qualitative examples of edge enhancement in our approach applied to both the conditional and unconditional model of Zhang \etal, and conditional of Su \etal~\cite{su2020insta}
In Fig.~\ref{fig:qual_3}, colorized images have the color-bleeding region for the baseline models. 
Comparing with other approaches, our method refines the coarse region.  
We present additional qualitative results of our approach with these two baselines in Figs.~\ref{fig:qual2} and \ref{fig:qual3}.

\begin{figure*}
\begin{center}
  \includegraphics[width=\linewidth]{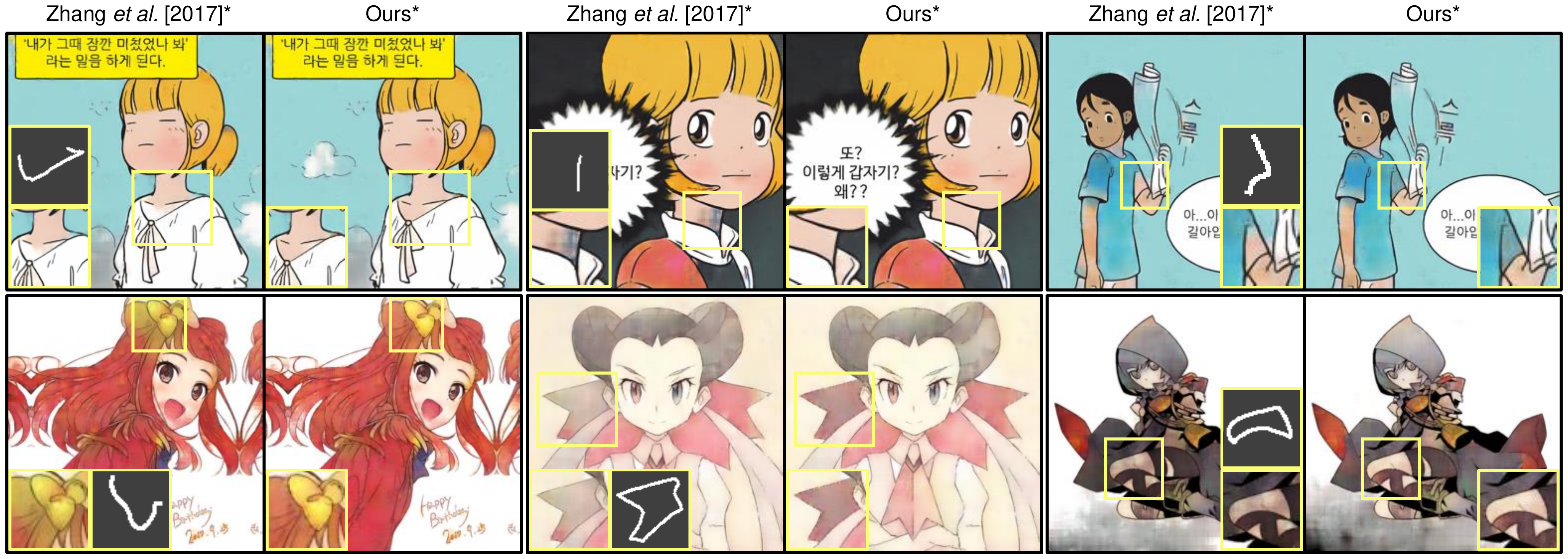}
  \end{center}
  \vspace{-0.6 cm}
  \caption{Qualitative results of edge enhancement in sketch colorization. We trained and evaluated our method on two datasets, \textit{Yumi's Cells}~\cite{yumicells} (first row) and \textit{Danbooru}~\cite{danbooru2017} (second row). For each row, the columns denoted as ``Zhang \etal" include the colorized outputs with color-bleeding artifacts, and the columns of ``ours" represent the edge-enhanced results.}
\label{fig:sketch_cb}
\vspace{-0.4cm}
\end{figure*}

\subsection{Quantitative Comparison}
\label{sec:quantitative_results}
As the improvement in the colorization outputs induced by our approach arises in a local region along the edges, we present the results of PSNR and LPIPS measured in these particular locations, as well as in a global region.
To conduct a local evaluation, we randomly sample $1,000$ $S_{\text{pseudo}}$ from the test samples of each dataset, which contain the regions of color-bleeding, as explained in Section~\ref{Pseudo-Scribble}.
Afterward, we report the local scores measured within the kernel size of $7$ along the edge pixels of those $S_{\text{pseudo}}$.
For global evaluation, we report the scores calculated on the entire region, specifying a kernel size as \textit{Full}.
To accommodate the scribbles similar to the real hints in our evaluation, we provide $S_{\text{pseudo}}$ with random widths ranging from 1 to 5 pixels for edge enhancement.
Table~\ref{tab:uncon_comp} shows that our model effectively advances the colorization performance of two baselines, outperforming the existing methods.
As bleeding effects are mostly mitigated in the local regions, large improvements are observed when the evaluation regions are localized along the edges.
In addition, Table~\ref{tab:cd_ratio} presents the CDR measured for both our method and the baselines.
Our method achieves a higher score against other baselines, demonstrating its superiority in colorizing the adjacent instances with different colors.

In Table~\ref{tab:ablation}, we ablate $\mathcal{L}_{\text{con}}$, $\mathcal{L}_{\text{reg}}$, and width augmentation technique to analyze their effectiveness quantitatively.
When we ablate $\mathcal{L}_{\text{con}}$, overall performance on PSNR, LPIPS, and CDR is slightly degraded, which is mainly due to the color distortion in the wrong regions.
Our method without $\mathcal{L}_{\text{reg}}$ results in degraded scores of PSNR and LPIPS while obtaining the best score in CDR.
Since $\mathcal{L}_{\text{reg}}$ suppresses the excessive perturbations in the refined feature maps of our edge-enhancing network, removing this loss causes an excessive edge enhancement (\textit{e.g.}, saturated colors along the edge) as well as undesirable color distortions in the entire region.
As we ablate the width augmentation for the $S_{\text{pseudo}}$ in the training, our approach achieves the best score in CDR, while PSNR and LPIPS score become even worse than Zhang \etal.
This implies that our edge-enhancing network tends to excessively enhance the color boundaries when unseen thick scribble is provided in the test time, ruining the overall colorization quality.
Therefore, width augmentation plays a critical role in learning a robust color enhancing, invariant to the width of a given scribble.
We support this claim with the qualitative results of the models with these objective functions ablated in Section~\ref{sec:ablation}.
In summary, our proposed method with every proposed objective function and augmented $S_{\text{pseudo}}$ achieves an optimal performance of edge enhancement.

\subsection{Verification for Labor-Efficient Interaction}
\label{scalability}
We believe that our approach provides fast and reliable interactions, which help users to correct the color bleeding in an intuitive manner.
To verify its usefulness with regard to labor efficiency, we conduct a user study with 13 novice participants using a user interface that we provide.
Each participant is given five randomly selected colorized images that contain the color-bleeding artifacts.
They are instructed to enhance the images by identifying color-bleeding areas and drawing scribbles until they obtain satisfactory results.
On average, finding color-bleeding areas and drawing scribbles take $13.20\pm8.46$ and $4.41\pm3.08$ seconds per image, and each participant draws $2.9\pm1.2$ scribbles to enhance an image.
The resultant improvement of PSNR is from $23.7$ to $26.2$ and LPIPS from $0.14$ to $0.07$, indicating that participants provide meaningful edges for enhancement.
These results demonstrate that our method have potentials to be applied in practical applications.
We provide additional analysis on the robustness of our method across different users in Section~\ref{sec:variance}.

\section{Experiments on Sketch Colorization}
\label{sec:05-sketch_colorization}
In this section, we further explore the potentials of our method in sketch colorization as well.
Compared to the gray-scale image, the sketch image contains a set of thin lines that explicitly define the semantic boundaries between objects.
However, as shown in the Fig.~\ref{fig:sketch_cb}, color-bleeding artifacts across these lines are easily observed, which indicates that the model still fails to preserve the boundary even when they contain edges in the input image.
The qualitative results in the columns denoted as ``ours" in Fig.~\ref{fig:sketch_cb} demonstrate that our method performs robust edge preservation in the sketch colorization as well.
We train and evaluate our method and the baselines on comic domain dataset including \textit{Yumi's Cells}~\cite{yumicells} and \textit{Danbooru}~\cite{danbooru2017}. 
The implementation details about sketch colorization are described in Section~\ref{sec:implementation_details}.
To further demonstrate the effectiveness of our method on sketch colorization, we present quantitative and qualitative results in \ref{sec:sketch_colorization}.

\vspace{0.2cm}
\begin{table}
\begin{center}
\begin{tabular}{@{}lclllclll@{}}
\toprule
\multicolumn{1}{c}{\multirow{2}{*}{Methods}} \phantom{ab} &
\multicolumn{3}{c}{ImageNet ctest~\cite{imagenet_cvpr09}} &\\
\cmidrule{2-4} 
& \multicolumn{1}{c}{LPIPS$\downarrow$} & \multicolumn{1}{c}{PSNR$\uparrow$} &
\multicolumn{1}{c}{CDR$\uparrow$}\\
\midrule
\multicolumn{1}{c}{Zhang \etal~\cite{zhang2017real}$^*$}& 0.208 & 14.966 & 0.418 \\
% \multicolumn{1}{l}{${\text{+Ours w/o } \mathcal{L}_{\text{edge}}}^*$} & 0.208 & 14.966 & 0.449\\ 
\multicolumn{1}{l}{${\text{+Ours w/o } \mathcal{L}_{\text{con}}}^*$} & 0.183 & 15.799 & 0.472\\ 
\multicolumn{1}{l}{${\text{+Ours w/o } \mathcal{L}_{\text{reg}}}^*$} & 0.185 & 15.624 & 0.605\\ 
\multicolumn{1}{l}{${\text{+Ours w/o aug }}^*$} & 0.259 & 13.372 & \textbf{0.647}\\
\multicolumn{1}{l}{${\text{+Ours (Full) }}^*$} & \textbf{0.177} & \textbf{16.041} & 0.512\\
\bottomrule
\end{tabular}
\end{center}
\vspace{-0.4cm}
\caption{Ablation study on the proposed modules. All results are measured within kernel size of 7 along with the single scribble.}
\label{tab:ablation}
\vspace{-0.4cm}
\end{table}

\section{Conclusion}
\label{sec:06-conclusion}
In this paper, we propose a novel and simple approach to effectively alleviate the color-bleeding artifacts which significantly degrades the quality of colorized outputs. Our method improves the bleeding edges by refining the intermediate features of the colorization network in the desired regions via user-interactive scribbles as additional inputs. Extensive experiments demonstrate its outstanding performance and reasonable labor efficiency, manifesting its potentials in practical applications.

\vspace{0.2cm}
\noindent 
\textbf{Acknowledgements} This work was supported by Institute of Information \& communications Technology Planning \& Evaluation (IITP) grant funded by the Korea government(MSIT) (No. 2019-0-00075, Artificial Intelligence Graduate School Program(KAIST)).
This work was also supported by Institute of Information \& communications Technology Planning \& Evaluation(IITP) grant funded by the Korea government(MSIT) (No. 2021-0-01778, Development of human image synthesis and discrimination technology below the perceptual threshold)
We thank all researchers at NAVER WEBTOON Corp.

{\small
\bibliographystyle{unsrt}
\bibliography{egbib}
}
\clearpage

\appendix
This supplementary material complements our paper with additional experimental results and their analysis. 
First, Section~\ref{sec:variance} presents how insensitive our edge-enhancing method is across the different real-world users.
Then, qualitative results of the ablation study are presented in Section~\ref{sec:ablation}, followed by an explanation about the analysis on our proposed metric CDR, with an example in Section~\ref{sec:discrepancy_ratio}.
In Section~\ref{sec:layers}, we provide the qualitative comparisons between our approach applied in encoder and decoder layers as well as their analysis.
As an interactive colorization approach, we construct our own user interface, where users can draw scribbles with adjustable widths for edge enhancement.
Section~\ref{sec:demo} describes this tool with the step-by-step demonstration.
Furthermore, we provide additional quantitative and qualitative results of our method applied to two baselines (Zhang \etal~\cite{zhang2017real} and Su \etal~\cite{su2020insta}) over various datasets, in Sections~\ref{sec:quan} and~\ref{sec:qual}.
Moreover, Section~\ref{sec:sketch_colorization} contains both quantitative results and qualitative examples of edge enhancements in sketch colorization, complementary to Section \ref{sec:05-sketch_colorization}.
Lastly, Section~\ref{sec:implementation_details} provides the implementation details, such as the settings for training, network architecture, and hyper-parameters for edge-extracting modules in the generation of $S_{\text{pseudo}}$.

\section{Variance of Edge Enhancement across Users}
\label{sec:variance}
\begin{figure}[t!]
\begin{center}
  \includegraphics[width=\linewidth]{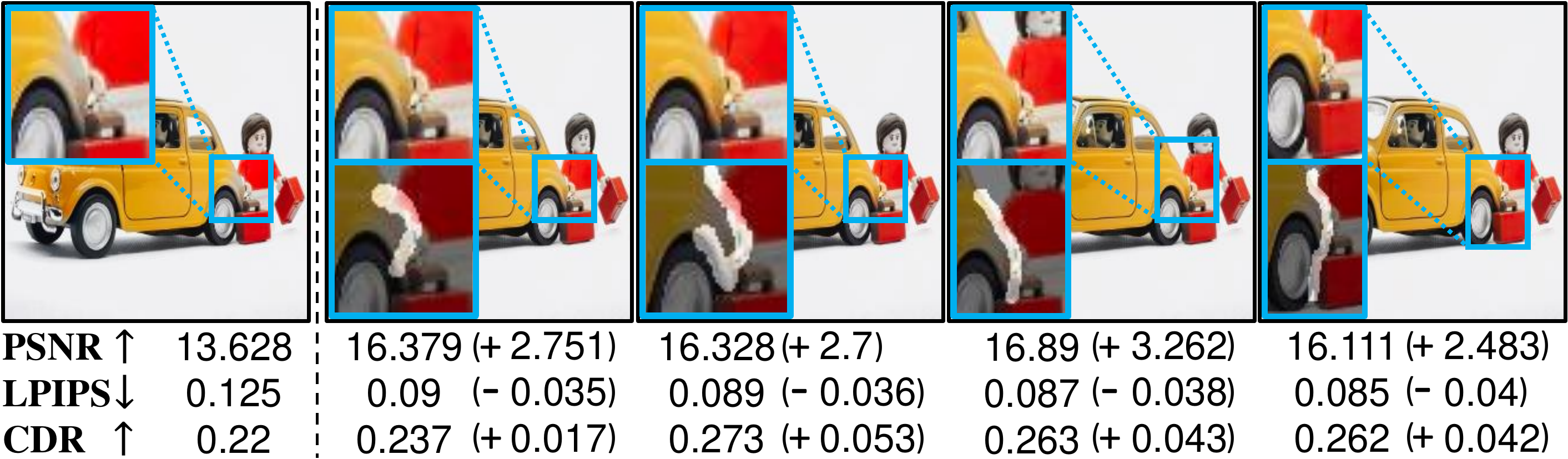}
  \end{center}
  \vspace{-0.4 cm}
  \caption{We visualize the initial colorization (first column) and its enhanced outputs by four different users (second to fifth column). 
  Enlarged views of edge-enhanced regions and user-given scribbles are provided in blue boxes. 
  Values in the brackets indicate the improved metric scores from those of the initial output. Zoom in for detail.}
  \vspace{-0.4 cm}
\label{fig:userstudy_sensitivity}
\end{figure}

To verify the robustness of the proposed method against ambiguous scribbles across different users, we measure the improved PSNR, LPIPS, and CDR on the enhanced colorization outputs obtained by different users.
To this end, among the color-bleeding edges that were used in the user study (Section \ref{scalability}), we selected the scribbles for the edges that were enhanced by at least four different participants in our user study.
For each edge, we calculated the improvement of these evaluation scores in the local regions near the scribbles drawn by the different users, following the same evaluation procedure in Section \ref{scalability}.
The resulting \textit{mean} and \textit{standard deviation} are $3.380\pm 1.381$, $0.031\pm 0.009$, and $0.044\pm 0.031$ for PSNR, LPIPS and CDR, respectively.
These results indicate that our method achieves consistent improvement under various scribbles drawn by different users.
In addition, we provide a qualitative example of these results in Fig.~\ref{fig:userstudy_sensitivity} with their evaluation scores.
Our method robustly improves the color-bleeding edges given varying styles of the user scribbles.

\section{Ablation Studies on Width Augmentation and Consistency Loss}
\label{sec:ablation}

\begin{figure*}[h]
\begin{center}
  \includegraphics[width=\linewidth]{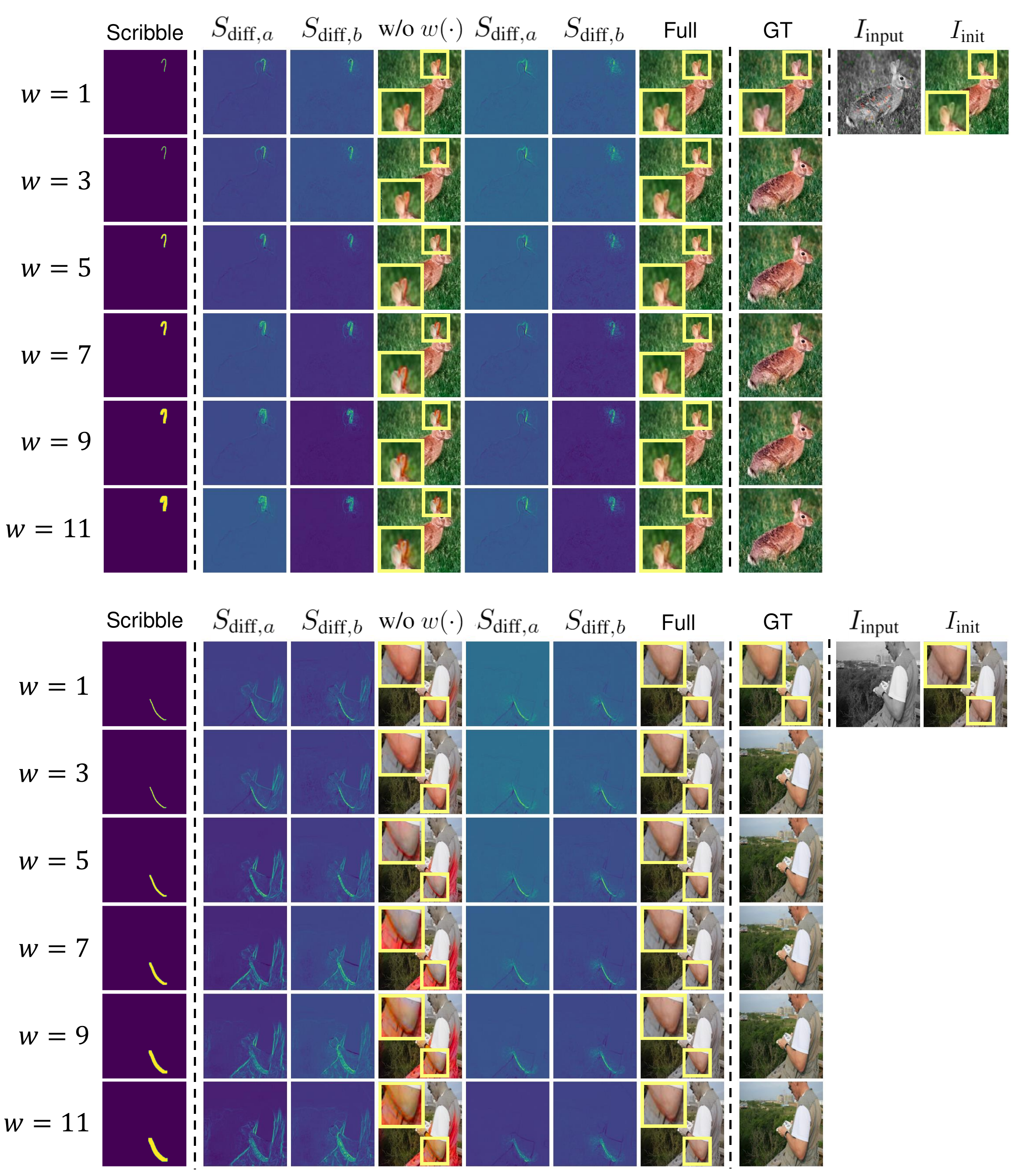}
  \end{center}
  \vspace{-0.6 cm}
  \caption{Qualitative comparisons between the generated outputs from our model trained with and without width augmentation $w(\cdot)$ for $S_{pseudo}$ in the training. 
  The fourth column contains the results of our model without $w(\cdot)$, and seventh column corresponds to the results of our model with all the proposed techniques, including $w(\cdot)$.}
\label{fig:sup_qual_ablation}
\end{figure*}

\begin{figure}[h]
\begin{center}
  \includegraphics[width=\linewidth]{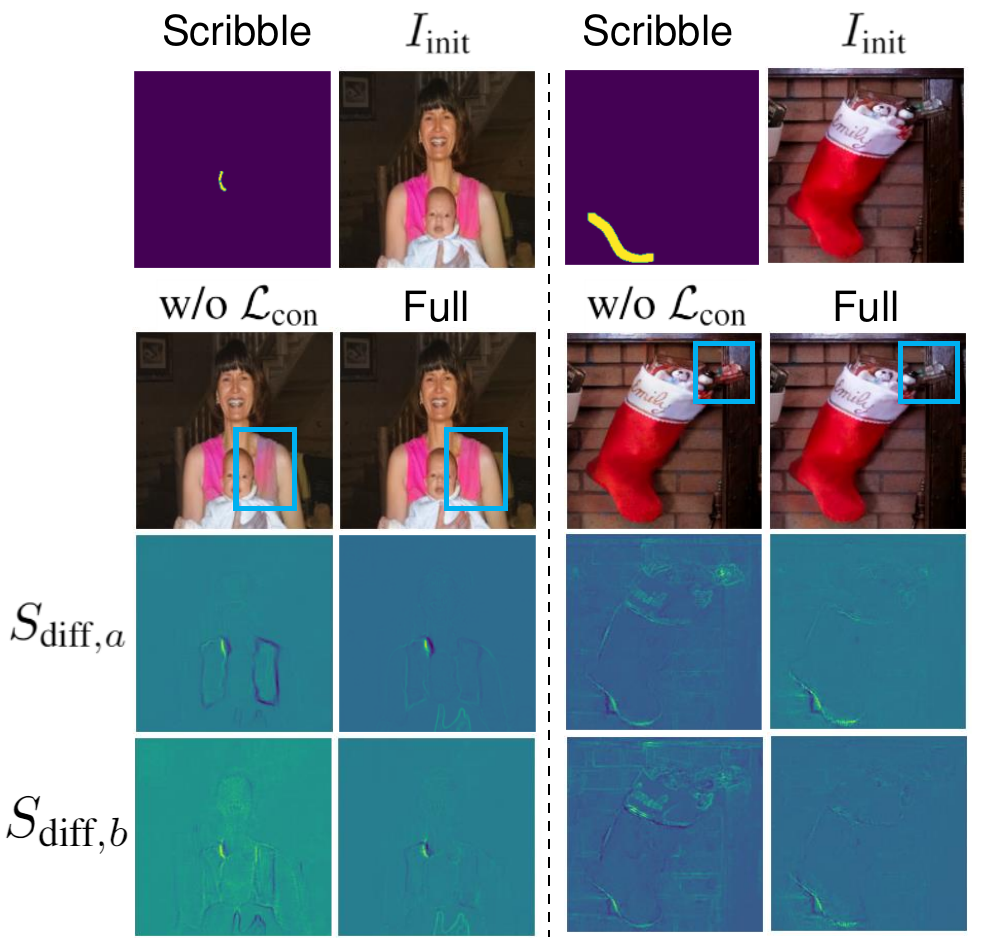}
  \end{center}
  \vspace{-0.6 cm}
  \caption{Qualitative comparisons between the generated outputs from our approach trained with and without $\mathcal{L}_{\text{con}}$ in the training.
  The column denoted as $\text{w/o} \ \mathcal{L}_{\text{con}}$ contains the results of our model without $\mathcal{L}_{\text{con}}$, and the column denoted as $\text{Full}$ corresponds to the results of our model with all the proposed techniques, including $\mathcal{L}_{\text{con}}$.}
  \vspace{-0.2 cm}
\label{fig:sup_qual_ablation2}
\end{figure}

This section provides the qualitative results of ablation studies on the scribble width augmentation $w(\cdot)$ and the consistency loss $\mathcal{L}_{\text{con}}$, described in the Section \ref{Pseudo-Scribble} and \ref{sec:objective_functions}, respectively.
In Fig.~\ref{fig:sup_qual_ablation}, the first column shows a binary map of user-driven scribble with its width varying from 1 to 11 pixels.
The columns named $S_{\text{diff}, a}$ and $S_{\text{diff}, b}$ represent how the edge values are changed in \textit{CIE ab} channels. 
$S_{\text{diff}, a}$ and $S_{\text{diff}, b}$ are obtained by subtracting $\mathcal{S}(I_{\text{init},ab})$ from $\mathcal{S}(I_{ab})$, where $\mathcal{S}(\cdot)$ approximates the edges, as described in Eq.~\ref{eq1}, $I_{\text{init},ab}$ is an initial colorized output by a backbone network and $I_{ab}$ is a refined output by our method.

The fourth column (w/o $\text{Aug}$) presents the generated outputs of the model trained without width augmentation in a training phase, while the seventh column (\text{Full}) contains the outputs with augmentation.
As the width of scribbles $\text{w}$ increases, especially when $\text{w}$ becomes larger than $5$, the width of changed edges $S_{\text{diff}, a}$ and $S_{\text{diff}, b}$ become thick as well.
This results in an excessive increase of edges in $ab$ channels, producing extremely vivid colors (\textit{e.g.}, red) along the given boundaries.
In contrast, our model trained with augmentations $w(\cdot)$  maintains the increase of enhanced edges regardless of given scribble's widths, robustly generating the plausible color corrections for all possible scribbles.
This allows the users to provide their interactions without giving much effort to drawing sharp thin scribbles.

Fig.~\ref{fig:sup_qual_ablation2} compares the qualitative results of our model trained with and without $\mathcal{L}_{\text{con}}$.
As explained in Section \ref{sec:objective_functions},  $\mathcal{L}_{\text{con}}$ enforces our model not to generate unintentional color changes outside of the target edges.
Therefore, the model trained without this objective function tends to produce unnecessary changes of colors in the wrong regions.
As shown in third and fourth rows, $S_{\text{diff}, a}$ and $S_{\text{diff}, b}$ of the model with $\mathcal{L}_{\text{con}}$ shows more sparse changes of pixels, compared to that without $\mathcal{L}_{\text{con}}$.
Ablation of $\mathcal{L}_{\text{con}}$ causes color distortions, such as washed-out colors or unintended color changes, illustrated in the images with blue bounding boxes.
For example, a woman with a pink vest contains a washed-out color on her right side of the vest.
This can be observed in the $S_{\text{diff}, a}$, specifically in the dark boundaries of the vest (decreased edges).

\section{Analysis on Cluster Discrepancy Ratio}
\label{sec:discrepancy_ratio}
\begin{figure}[h]
\begin{center}
  \includegraphics[width=\linewidth]{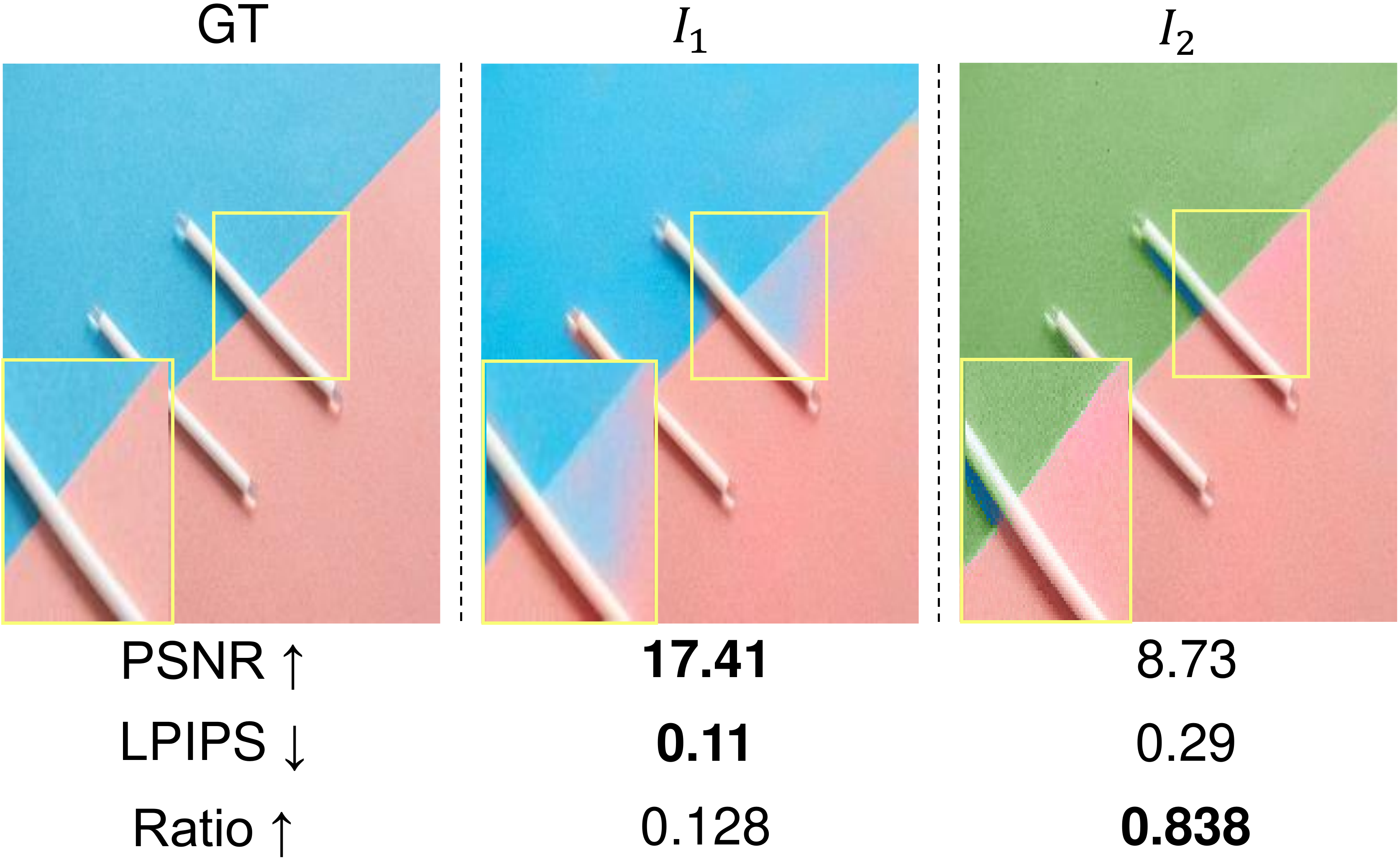}
  \end{center}
  \vspace{-0.2 cm}
  \caption{Quantitative comparisons of the colorized outputs, $I_1$ and $I_2$, via PSNR, LPIPS~\cite{zhang2018perceptual} and CDR.
  The yellow box provides the enlarged view on the regions of color-bleeding effects appearing in $I_1$, while enhanced in $I_2$ by our proposed method.
  Three evaluation scores for each image are presented below.}
\label{fig:sup_discrepancy_ratio}
\vspace{-0.4 cm}
\end{figure}

This section provides a specific example that demonstrates the necessity of our proposed metric, \textit{i.e.}, CDR.
As mentioned in Section~\ref{sec:ratio}, this metric aims to cover the blind spot of two generally used metrics PSNR and LPIPS~\cite{zhang2018perceptual} in the colorization task.
In Fig.~\ref{fig:sup_discrepancy_ratio}, the first and the second columns are the ground-truth and colorized output $I_1$ with color-bleeding artifacts in the yellow box, respectively.
$I_2$ is another example of colorized output with its bleeding edge enhanced by our proposed approach.
Note that this is colorized with a different color from the ground-truth, enabled by providing user-interactive color hints.
As $I_2$ contains the different colors from the ground-truth, unlike $I_1$, it is shown to record lower PSNR and higher LPIPS score than $I_1$.
However, $I_2$ contains a \textit{clear color edge}, while $I_1$ contains the color-bleeding effects, which make $I_2$ more visually favorable than $I_1$.
The results of these metrics on $I_2$ against $I_1$ represent that they often underrate the quality of realistic sharp images that contain different colors from ground-truth.
Our metric, however, essentially evaluates whether the colors are different across the adjacent objects and less dependent on the ground-truth colors.
Therefore, the discrepancy ratio achieves a higher score for $I_2$ compared to $I_1$, showing that it is possible to robustly evaluate the color-bleeding artifacts.

Note that the sole use of our metric for evaluating colorization methods also reveals a bottleneck as mentioned in Section \ref{sec:quantitative_results}.
More specifically, as our metric focuses on evaluating whether the colors are different between edges, saturated colors along the edges can be evaluated as favorable.
Therefore, it is recommended to consider all of these metrics to fully evaluate the general colorization performance, including the edge preservation.

\begin{figure}[b]
\begin{center}
  \includegraphics[width=\linewidth]{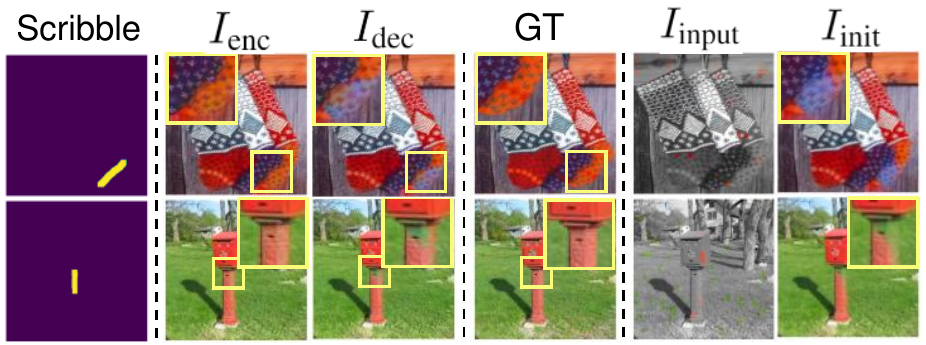}
  \end{center}
  \vspace{-0.6 cm}
  \caption{Qualitative comparisons of edge-enhancing network applied in encoder and decoder layers. $I_{\text{enc}}$ and $I_{\text{dec}}$ denote the enhanced results of the respective setting.}
\label{fig:sup_enc_dec}
\end{figure}

\section{Selecting Layers for Edge-Enhancing Network}
\label{sec:layers}
As briefly described in Section \ref{edge_refinement_network}, we empirically find that applying edge-enhancing network $E$ in the encoder layers achieves the intended results.
To show the effectiveness of this choice of layers, Fig.~\ref{fig:sup_enc_dec} compares the results of $E$ applied in encoder and decoder layers, $I_{\text{enc}}$ and $I_{\text{dec}}$.
It is shown that $I_{\text{enc}}$ successfully preserves the edges between the different objects, correcting the colors spreading in the regions.
For example, by giving a vertical scribble, wrongly spread green pixels inside the fire hydrant are corrected to a red color.
However, $I_{\text{dec}}$ merely increases the color difference along the edges without removing the green pixels inside the edges.
We believe that $E$ applied in the encoder layers refines the representations related to the edges, helping the following layers to spread the colors corresponding to the enhanced edges.
In this regard, we choose the encoder layers of the backbone network as an appropriate option for integrating $E$.

\begin{figure*}
\begin{center}
  \includegraphics[width=\linewidth]{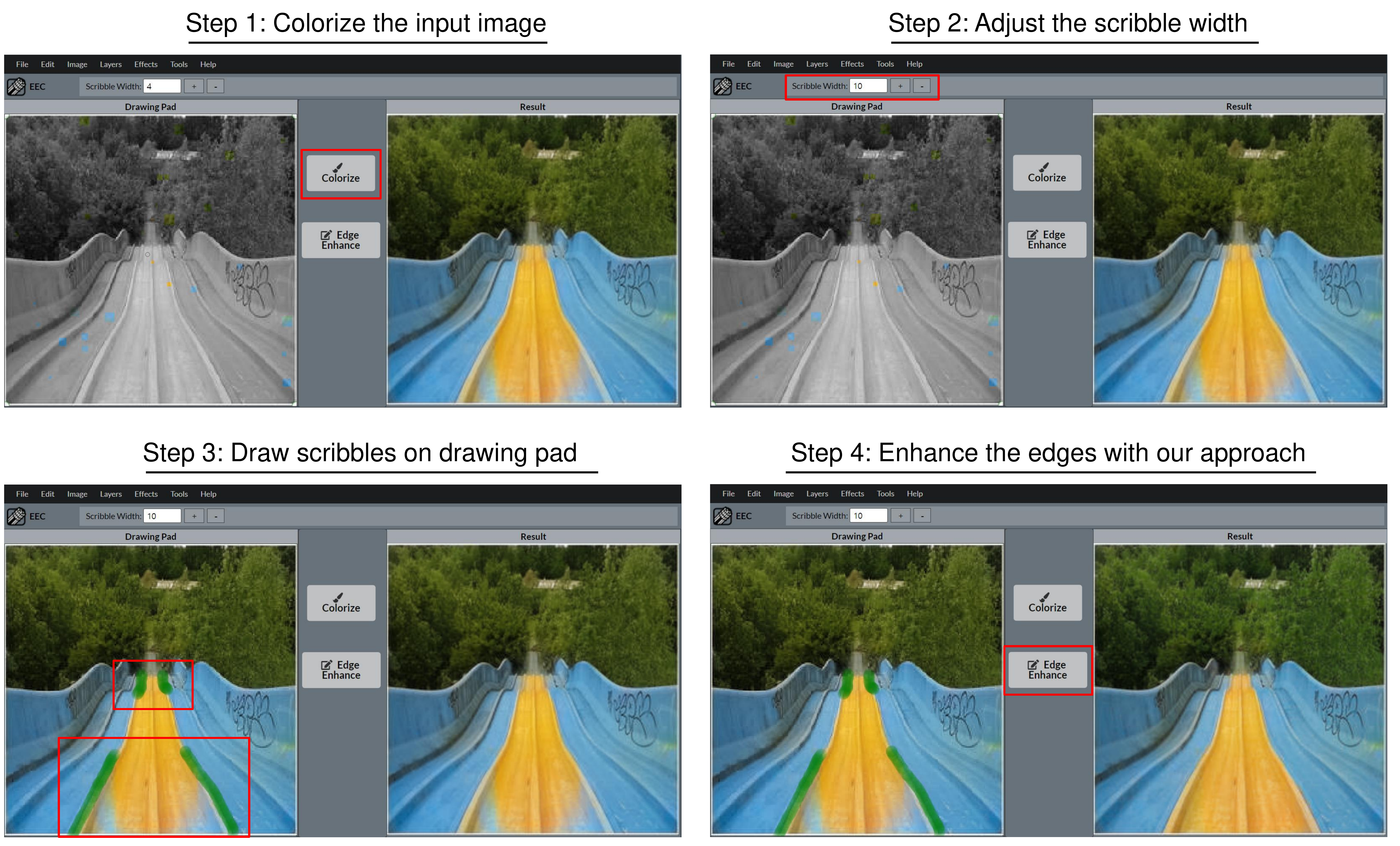}
  \end{center}
  \vspace{-0.6 cm}
  \caption{A step-by-step demonstration of an edge enhancement via our proposed user interface tool.}
\label{fig:demo}
\end{figure*}

\section{User-Interaction Demonstration}
\label{sec:demo}
This section provides a step-by-step demonstration of edge enhancement through our user interface tool, as illustrated in Fig.~\ref{fig:demo}.
First, the user uploads and colorizes an input image (left panel) with additional color hints by clicking the \textit{Colorize} button.
Then, the user adjusts the width of the scribble and draws on color-bleeding edges of the colorized image shown in the left panel.
After applying the scribbles, clicking the \textit{Edge Enhance} button forwards the drawn scribbles into our edge-enhancing network to correct the annotated edges and show the improved result in the right panel.

\section{Additional Quantitative Results}
\label{sec:quan}
As described in Section~\ref{sec:04-experiments}, we apply CIC~\cite{zhang2016colorful}, DeOldify~\cite{deoldify} and Zhang \etal~\cite{zhang2017real} as our baseline models for unconditional colorization, and Zhang \etal~\cite{zhang2017real}, Su \etal~\cite{su2020insta} as our baseline for conditional colorization task.
Table~\ref{tab:kernel_test} contains the additional rows of quantitative comparisons of our model and baselines in the kernel size of $15$ and $23$, complementary to Table \ref{tab:uncon_comp}.
For the qualitative results, we demonstrate that our framework is applicable to Zhang\etal, which is the most widely used conditional colorization method, and Su \etal, which is a recently proposed colorization approach.
Utilizing our approach on these backbone networks improves the general colorization quality over the various dataset, outperforming the existing baselines.
As our approach significantly improves the local color-bleeding artifacts, it improves the applied baselines especially when evaluated near the edge regions (\textit{i.e.}, small kernel size).

Table~\ref{tab:Width} demonstrates the robustness of edge enhancing performance applied to Zhang \etal when the width of a given scribble varies from $1$ to $9$ pixel diameter.
For the CDR, we evaluate it within the kernel size of $7$ along the given edges.
We observe that our model robustly enhances the local colorization performance in every dataset, given any scribble widths.
Similar to the Table~\ref{tab:kernel_test}, the difference of global score between our model and baselines becomes minor when averaging the scores over all the spatial dimensions.

\begin{table*}
\begin{center}
\begin{tabular}{@{}lclllclllclll@{}}
\toprule
\multicolumn{1}{c}{\multirow{2}{*}{Kernel Size}} & \multicolumn{1}{c}{\multirow{2}{*}{Methods}} \phantom{abc} &
\multicolumn{2}{c}{ImageNet ctest~\cite{imagenet_cvpr09}} & \phantom{ab} &
\multicolumn{2}{c}{COCO-Stuff~\cite{caesar2018coco}} & \phantom{ab} &
\multicolumn{2}{c}{Place205~\cite{zhou2017places}} & \\
\cmidrule{3-4} \cmidrule{6-7} \cmidrule{9-10}
&& \multicolumn{1}{c}{LPIPS$\downarrow$} & \multicolumn{1}{c}{PSNR$\uparrow$} && \multicolumn{1}{c}{LPIPS$\downarrow$} & \multicolumn{1}{c}{PSNR$\uparrow$} && \multicolumn{1}{c}{LPIPS$\downarrow$} & \multicolumn{1}{c}{PSNR$\uparrow$} \\
\midrule
\multicolumn{1}{c}{\multirow{8}{*}{K=7}}
& CIC ~\cite{zhang2016colorful} & 0.248 & 13.281 && 0.247 & 13.368 && 0.254 & 13.577\\
& DeOldify ~\cite{deoldify} & 0.250 & 13.234 && 0.251 & 13.059 && 0.227 & 14.258\\
& Zhang \etal~\cite{zhang2017real}& 0.246 & 13.248 && 0.206 & 14.755 && 0.219 & 14.815\\
& +Ours & \textbf{0.217}& \textbf{13.919} && \textbf{0.192} & \textbf{15.037} && \textbf{0.211} & \textbf{15.104}\\\cline{2-11}
& Zhang \etal~\cite{zhang2017real}$^*$& 0.208 & 14.966 && 0.158 & 17.456 && 0.171 & 17.530\\
& +Ours$^*$ & \textbf{0.177} & \textbf{16.041} && \textbf{0.143} & \textbf{17.953} && \textbf{0.161} & \textbf{17.906}\\\cline{2-11}
& Su \etal~\cite{su2020insta}$^*$& 0.185 & 16.393 && 0.187 & 15.971 && 0.194 & 17.032 \\
& +Ours$^*$ & \textbf{0.177} & \textbf{16.507} && \textbf{0.176} & \textbf{16.188} && \textbf{0.187} & \textbf{17.098}\\
\midrule
\multicolumn{1}{c}{\multirow{8}{*}{K=15}}
& CIC ~\cite{zhang2016colorful} & 0.221 & 14.377 && 0.237 & 13.937 && 0.219 & 14.841\\
& DeOldify ~\cite{deoldify} & 0.218 & 14.399 && 0.217 & 14.362 && 0.226 & 14.382\\
& Zhang \etal~\cite{zhang2017real} & 0.209 & 14.662 && 0.204 & 14.975 && 0.218 & 14.962\\
& +Ours & \textbf{0.195} & \textbf{14.963} && \textbf{0.191} &
\textbf{15.144} && \textbf{0.210} & \textbf{15.204}\\\cline{2-11}
& Zhang \etal~\cite{zhang2017real}$^*$& 0.181 & 16.278 && 0.155 & 17.708 &&  0.169 & 17.739 \\
& +Ours$^*$ & \textbf{0.159} & \textbf{17.113} && \textbf{0.142} & \textbf{18.144} && \textbf{0.161} & \textbf{18.054}\\\cline{2-11}
& Su \etal~\cite{su2020insta}$^*$& 0.166 & 17.335 && 0.169 & 17.000 && 0.177 & 17.889\\
& +Ours$^*$ & \textbf{0.159} & \textbf{17.477} && \textbf{0.159} & \textbf{17.225} && \textbf{0.170} & \textbf{18.004}\\
\midrule
\multicolumn{1}{c}{\multirow{8}{*}{K=23}}
& CIC ~\cite{zhang2016colorful} & 0.221 & 14.394 && 0.220 & 14.596 && 0.227 & 14.635\\
& DeOldify ~\cite{deoldify}& 0.226 & 14.171 && 0.216 & 14.431 && 0.225 & 14.458\\
& Zhang \etal~\cite{zhang2017real} & 0.210 & 14.696 && 0.204 & 14.975 && 0.217 & 15.056\\
& +Ours & \textbf{0.195} & \textbf{14.987} && \textbf{0.191} & \textbf{15.180} && \textbf{0.210} & \textbf{15.282}\\\cline{2-11}
& Zhang \etal~\cite{zhang2017real}$^*$& 0.164 & 17.195 && 0.154 & 17.806 && 0.169 & 17.864\\
& +Ours$^*$ & \textbf{0.147} & \textbf{17.868} && \textbf{0.141} & \textbf{18.204} && \textbf{0.161} & \textbf{18.147}\\\cline{2-11}
& Su \etal~\cite{su2020insta}$^*$& 0.154 & 17.945 && 0.156 & 17.611 && 0.164 & 18.567\\
& +Ours$^*$ & \textbf{0.148} & \textbf{18.081} && \textbf{0.148} & \textbf{17.840} && \textbf{0.158} & \textbf{18.689}\\
\midrule
\multicolumn{1}{c}{\multirow{8}{*}{K=Full}}
& CIC ~\cite{zhang2016colorful}& 0.172 & 21.001 && 0.164 & 21.456 && 0.153 & 21.873\\
& DeOldify ~\cite{deoldify} & 0.159 & 21.433 && 0.149 & 21.985 && 0.156 & 21.933\\
& Zhang \etal~\cite{zhang2017real} & 0.148 & 21.981 && 0.135 & 22.729 && 0.138 & \textbf{22.846}\\
& +Ours & \textbf{0.147} & \textbf{22.026} && \textbf{0.134} & 22.729 && 0.138 & 22.845\\\cline{2-11}
& Zhang \etal~\cite{zhang2017real}$^*$& 0.086 & 27.202 && 0.080 & 27.681 && 0.087 & 27.697\\
& +Ours $^*$& \textbf{0.085} & \textbf{27.559} && \textbf{0.078} & \textbf{27.955} && 0.087 & \textbf{27.935}\\\cline{2-11}
& Su \etal~\cite{su2020insta}$^*$& 0.091 & 26.211 && 0.089 & 26.050 && 0.090 & 27.414\\
& +Ours$^*$ & 0.091 & \textbf{26.291} && \textbf{0.088} & \textbf{26.233} && \textbf{0.089} & \textbf{27.486}\\
\bottomrule
\end{tabular}
\end{center}
\caption{Quantitative comparison with the baselines on 1,000 images in the ImageNet ctest~\cite{imagenet_cvpr09}, COCO-Stuff~\cite{caesar2018coco} and Place205~\cite{zhou2017places} validation set. Quantitative results in the local region show that our method effectively enhances the images.} 
\label{tab:kernel_test}
\end{table*}

\begin{table*}
\begin{center}
\begin{tabular}{ccccccccc}
\toprule
\multirow{2}{*}{Kernel Size} & 
\multirow{2}{*}{Datasets} & \multirow{2}{*}{Metrics} &
\multirow{2}{*}{\specialcell{Zhang \etal\\~\cite{zhang2017real}$^*$}} &
\multicolumn{5}{c}{Scribble Width} \\
\cmidrule{5-9}
&&& & 1 & 3 & 5 & 7 & 9 \\
\midrule
\multirow{9}{*}{K=7}
& \multirow{3}{*}{ImageNet ctest~\cite{imagenet_cvpr09}}
& LPIPS$\downarrow$ & 0.208 & 0.182 & 0.178 & 0.174 & \textbf{0.166} & 0.203 \\
&& PSNR$\uparrow$ & 14.966 & 15.877 & 16.040 & 16.022 & \textbf{16.303} & 14.862 \\
&& CDR$\uparrow$ & 0.376 & 0.411 & 0.436 & 0.451 & \textbf{0.471} & 0.447 \\\cline{3-9}
& \multirow{3}{*}{COCO-Stuff~\cite{caesar2018coco}}
& LPIPS$\downarrow$& 0.211 & 0.185 & 0.181 & 0.174 & \textbf{0.168} & 0.180 \\
&& PSNR$\uparrow$& 14.964 & 15.790 & 15.938 & 16.074 & \textbf{16.210} & 15.594 \\
&& CDR$\uparrow$& 0.372 & 0.417 & 0.446 & 0.465 & \textbf{0.478} & 0.459 \\\cline{3-9}
& \multirow{3}{*}{Place205~\cite{zhou2017places}}
& LPIPS$\downarrow$& 0.223 & 0.205 & 0.201 & 0.193 & \textbf{0.190} & 0.201 \\
&& PSNR$\uparrow$& 15.091 & 15.714 & 15.877 & 16.036 & \textbf{16.100} & 15.540 \\
&& CDR$\uparrow$& 0.330 & 0.389 & 0.409 & 0.431 & \textbf{0.438} & 0.425 \\
\midrule
\multirow{9}{*}{K=Full}
& \multirow{3}{*}{ImageNet ctest~\cite{imagenet_cvpr09}}
& LPIPS$\downarrow$& 0.086 & 0.085 & 0.085 & 0.085 & \textbf{0.084} & 0.087  \\
&& PSNR$\uparrow$& 27.202 & 27.555 & \textbf{27.558} & 27.510 & 27.554 & 27.293 \\
&& CDR$\uparrow$& \textendash & \textendash & \textendash & \textendash & \textendash & \textendash 
\\\cline{3-9}
& \multirow{3}{*}{COCO-Stuff~\cite{caesar2018coco}}
& LPIPS$\downarrow$& 0.080 & 0.078 & 0.078 & 0.078 & 0.078 & 0.079 \\
&& PSNR$\uparrow$& 27.677 & \textbf{27.964} & 27.959 & 27.953 & 27.939 & 27.880 \\
&& CDR$\uparrow$& \textendash & \textendash & \textendash & \textendash & \textendash & \textendash 
\\\cline{3-9}
& \multirow{3}{*}{Place205~\cite{zhou2017places}}
& LPIPS$\downarrow$& 0.087 & 0.087 & 0.087 & 0.087 & 0.087 & 0.087 \\
&& PSNR$\uparrow$& 27.697 & \textbf{27.945} & 27.931 & 27.926 & 27.897 & 27.877 \\
&& CDR$\uparrow$& \textendash & \textendash & \textendash & \textendash & \textendash & \textendash\\
\bottomrule
\end{tabular}
\end{center}
\caption{Quantitative results based on scribble width.
The scores for the CDR are reported only in the local regions along the edges, within the kernel size of 7.} 
\label{tab:Width}
\end{table*}

\section{Additional Qualitative Results}
\label{sec:qual}
\begin{figure*}
\begin{center}
  \includegraphics[width=\linewidth]{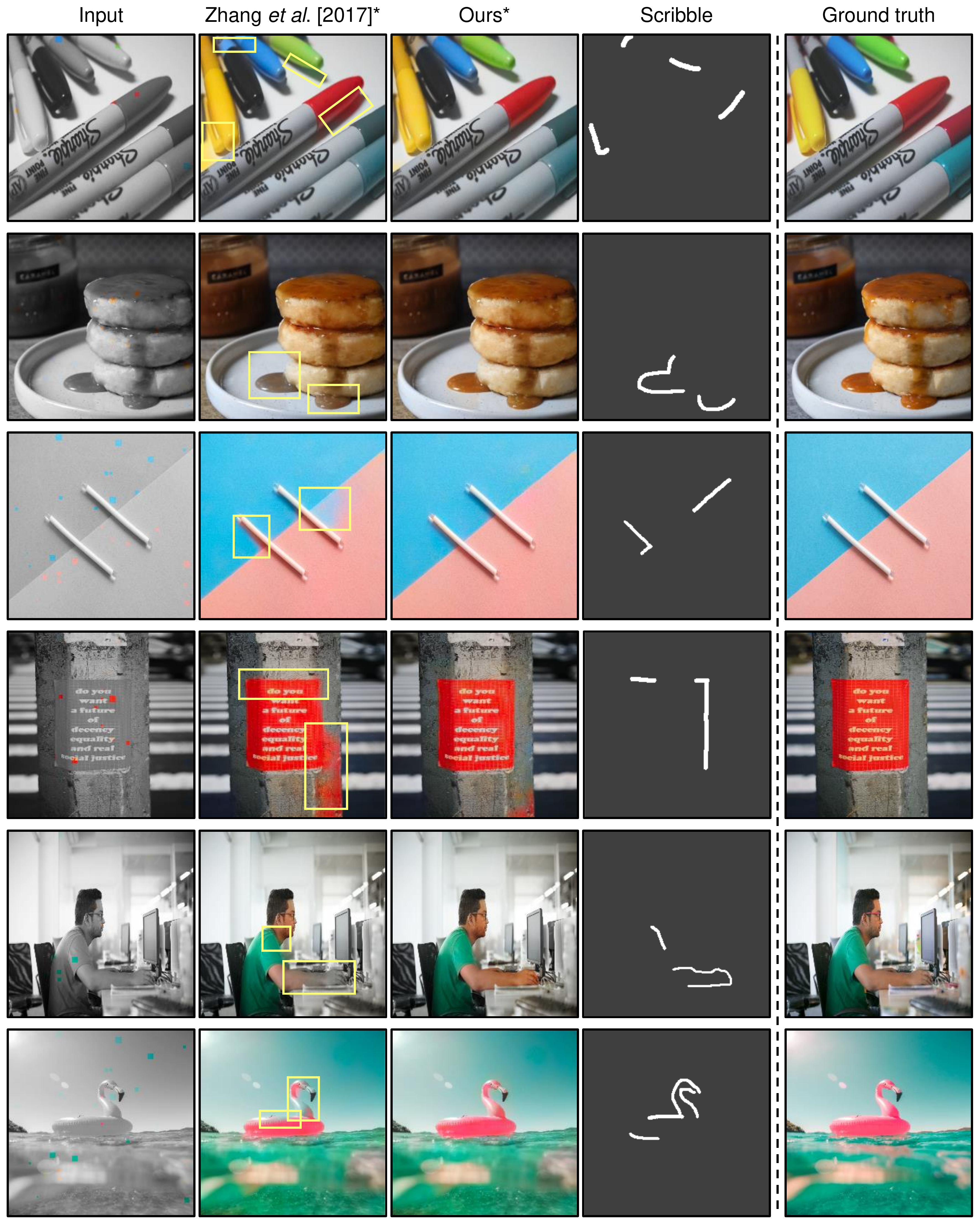}
  \end{center}
  \vspace{-0.6 cm}
  \caption{Qualitative results of edge-enhancement of our method applied to Zhang \etal~\cite{zhang2017real}}
\label{fig:qual2}
\end{figure*}

\begin{figure*}
\begin{center}
  \includegraphics[width=\linewidth]{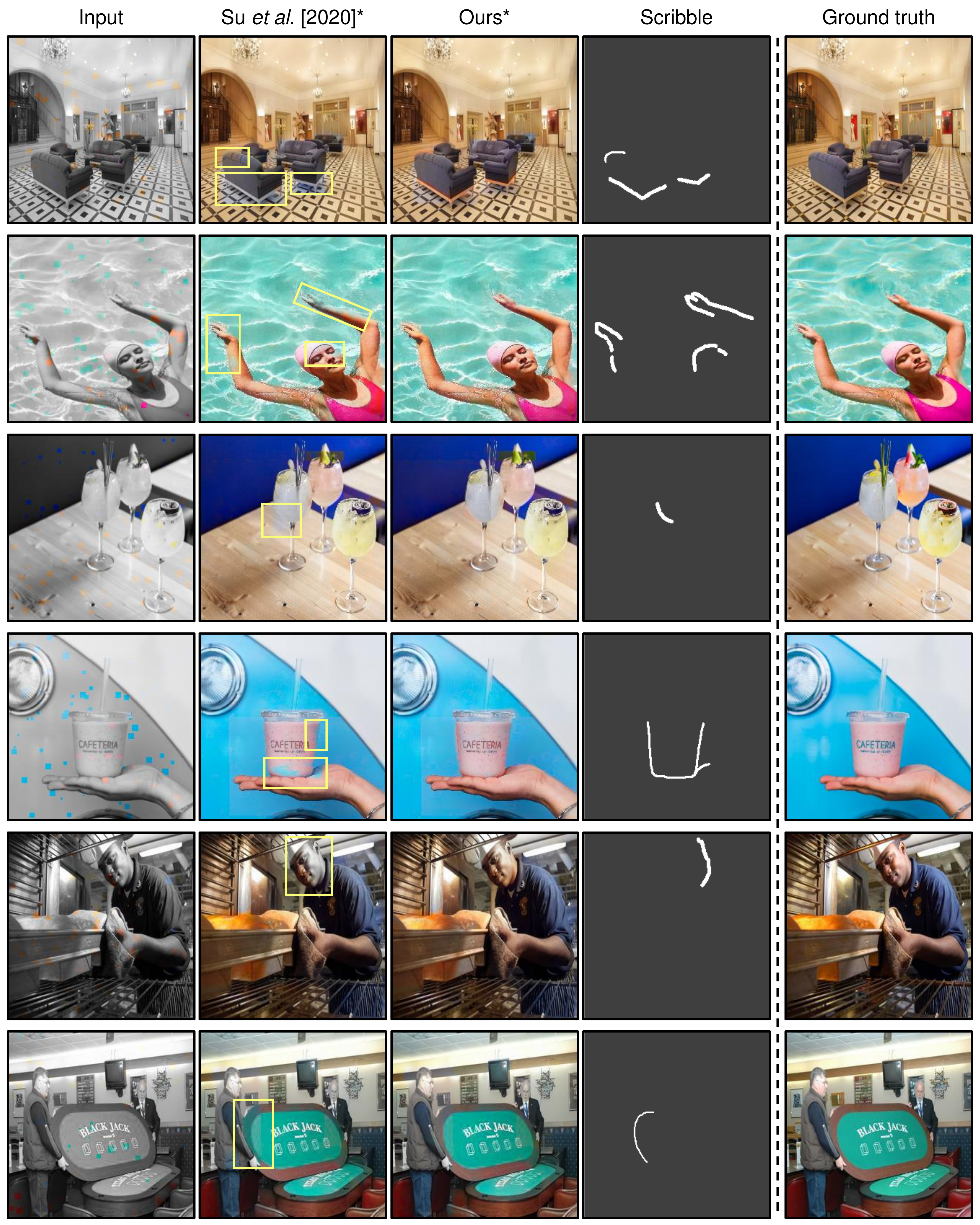}
  \end{center}
  \vspace{-0.6 cm}
  \caption{Qualitative results of edge-enhancement of our method applied to Su \etal~\cite{su2020insta}}
\label{fig:qual3}
\end{figure*}

Figs.~\ref{fig:qual2} and \ref{fig:qual3} present the qualitative examples of edge enhancement applied to Zhang \etal~\cite{zhang2017real}, and Su \etal~\cite{su2020insta}, respectively. 
The second and third columns of the figures contain the inference outputs of the baselines and their enhanced images using our method, respectively.
Their scribbles, which are used to enhance the images, are visualized in the fourth column. 
Yellow boxes in the second column represent color-bleeding areas.
All the images are collected from \url{https://unsplash.com} and Place205~\cite{zhou2017places}.

\begin{figure*}
\begin{center}
  \includegraphics[width=\linewidth]{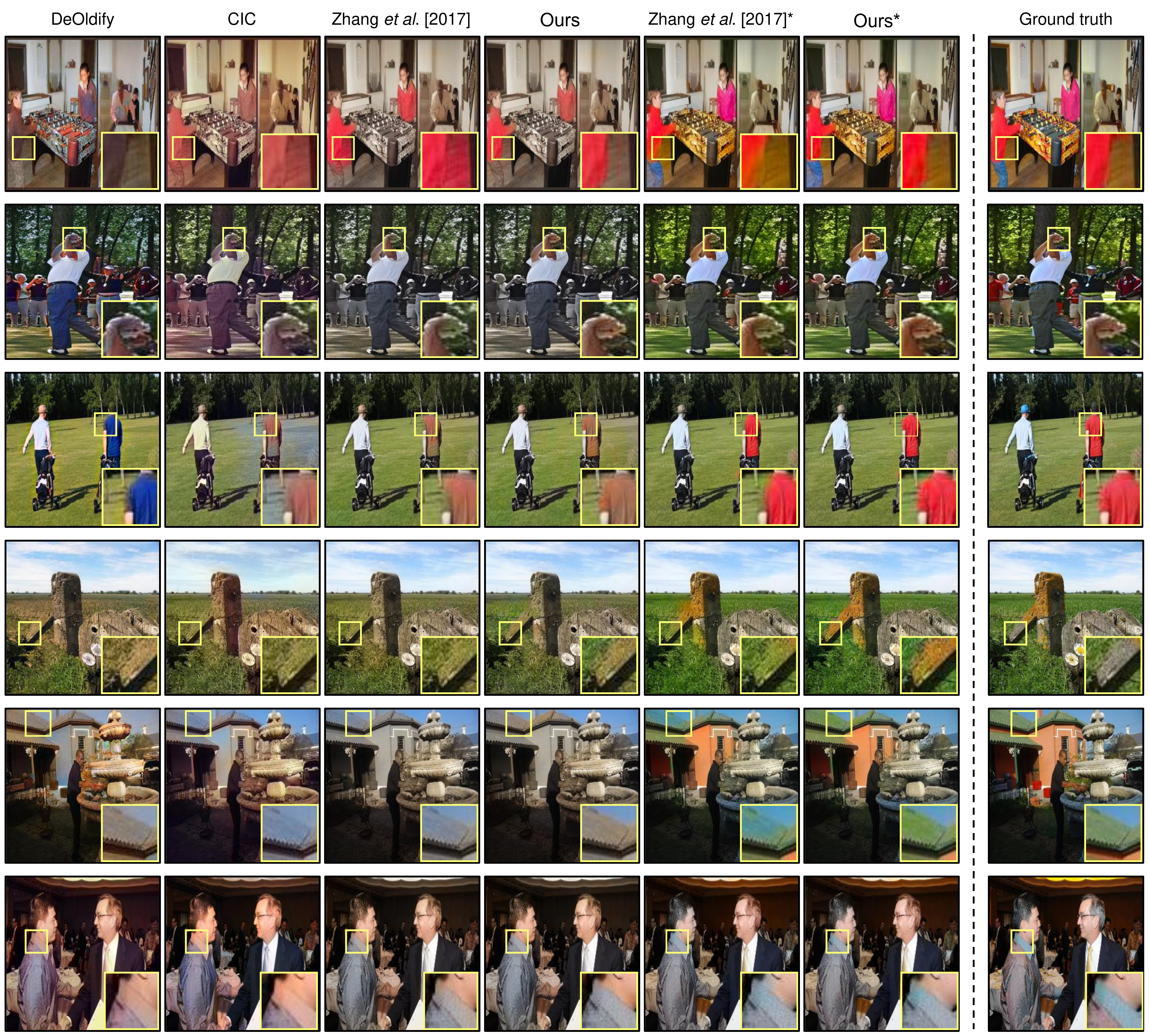}
  \end{center}
  \vspace{-0.6 cm}
  \caption{Qualitative comparisons between our model applied to Zhang \etal~\cite{zhang2017real} and other baseline models.}
\label{fig:qual1}
\end{figure*}

The qualitative comparisons between our method, especially applied to Zhang \etal~\cite{zhang2017real}, and other baselines, are presented in Fig.~\ref{fig:qual1}.
While all the baselines in the figure contain the color-bleeding artifact in the regions bounded with a yellow box, our approach improves the edges, as shown in the fourth and sixth columns.
Yellow boxes indicate the color-bleeding areas, and we provide the enlarged views of the areas on the right lower corner in the images. The images are selected from COCO-Stuff~\cite{caesar2018coco} and Place205~\cite{zhou2017places}.

\section{Sketch Colorization}
\label{sec:sketch_colorization}
\begin{figure*}
\begin{center}
  \includegraphics[width=\linewidth]{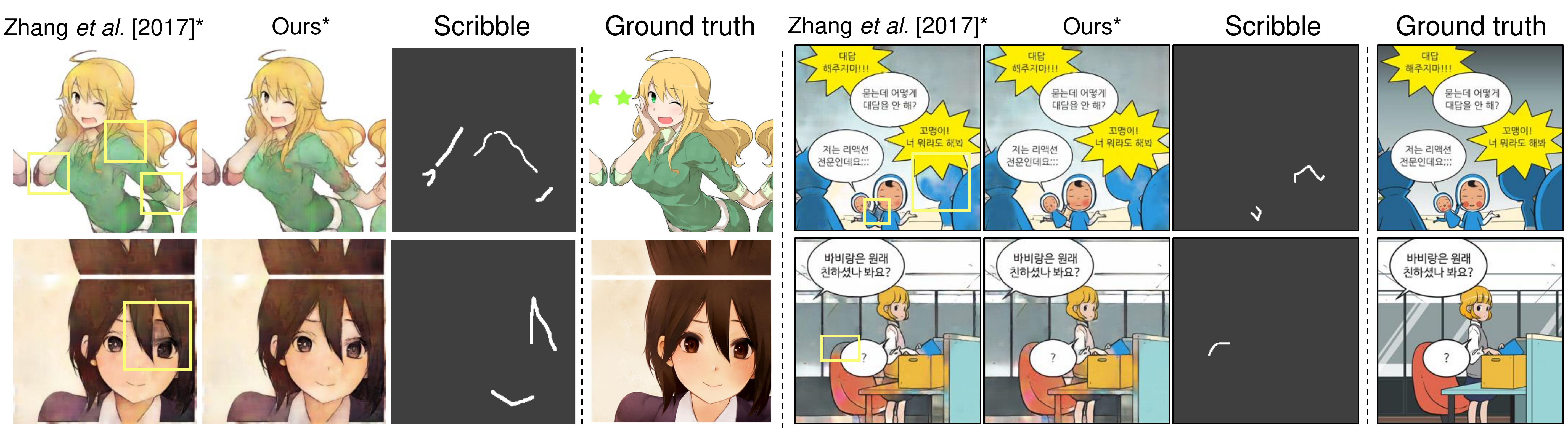}
  \end{center}
  \vspace{-0.6 cm}
  \caption{Qualitative examples of edge enhancement on sketch colorization. Yellow boxes provide a view of color-bleeding regions.}
\label{fig:sketch_qual}
\end{figure*}

In Section \ref{sec:05-sketch_colorization}, we demonstrate that our approach has the potential to enhance the edges in the sketch colorization task as well.
In this study, we utilize the two datasets, Yumi's Cells~\cite{yumicells} and Danbooru~\cite{danbooru2017}.
Table~\ref{tab:sketch_q} demonstrates that applying edge-enhancing network on the colorization model, newly trained for sketch colorization task, can also improve the performance, especially along the edge regions.
Additional qualitative results complementary to the Fig~\ref{fig:sketch_cb} are illustrated in Fig.~\ref{fig:sketch_qual}.

\begin{table*}
\begin{center}
\begin{tabular}{@{}lclllclllclll@{}}
\toprule
\multicolumn{1}{c}{\multirow{2}{*}{Kernel Size}} & \multicolumn{1}{c}{\multirow{2}{*}{Methods}} \phantom{abc} &
\multicolumn{3}{c}{Yumi's Cells~\cite{yumicells}} & \phantom{ab} &
\multicolumn{3}{c}{Danbooru~\cite{danbooru2017}} & \\
\cmidrule{3-5} \cmidrule{7-9} 
&& \multicolumn{1}{c}{LPIPS$\downarrow$} & \multicolumn{1}{c}{PSNR$\uparrow$} & CDR$\uparrow$ && \multicolumn{1}{c}{LPIPS$\downarrow$} & \multicolumn{1}{c}{PSNR$\uparrow$} & CDR$\uparrow$ \\
\midrule
\multicolumn{1}{c}{\multirow{2}{*}{K=7}}
& Zhang \etal~\cite{zhang2017real}$^*$& 0.240 & 10.738 & 0.231  && 0.322 & 11.970 & 0.191 \\
& +Ours$^*$ & \textbf{0.226} & \textbf{11.287} & \textbf{0.298} && \textbf{0.317} & \textbf{12.335} & \textbf{0.214} \\
\midrule
\multicolumn{1}{c}{\multirow{2}{*}{K=15}}
& Zhang \etal~\cite{zhang2017real}$^*$& 0.217 & 12.194 & \multicolumn{1}{c}{\textendash}  && 0.374 & 9.829 & \multicolumn{1}{c}{\textendash}   \\
& +Ours$^*$ & \textbf{0.210} & \textbf{12.487} & \multicolumn{1}{c}{\textendash}  && \textbf{0.371} & \textbf{10.092} & \multicolumn{1}{c}{\textendash} \\
\midrule
\multicolumn{1}{c}{\multirow{2}{*}{K=23}}
& Zhang \etal~\cite{zhang2017real}$^*$& 0.219 & 12.211 & \multicolumn{1}{c}{\textendash}  && 0.290 & 12.146 & \multicolumn{1}{c}{\textendash}   \\
& +Ours$^*$ & \textbf{0.211} & \textbf{12.531} & \multicolumn{1}{c}{\textendash}  && \textbf{0.286} & \textbf{12.448} & \multicolumn{1}{c}{\textendash} \\
\midrule
\multicolumn{1}{c}{\multirow{2}{*}{K=Full}}
& Zhang \etal~\cite{zhang2017real}$^*$& \textbf{0.153} & 19.280 & \multicolumn{1}{c}{\textendash}  && 0.204 & 18.698 & \multicolumn{1}{c}{\textendash}   \\
& +Ours$^*$ & 0.154 & \textbf{19.291} & \multicolumn{1}{c}{\textendash}  && \textbf{0.201} & \textbf{18.975} & \multicolumn{1}{c}{\textendash} \\
\bottomrule
\end{tabular}
\end{center}
\caption{Quantitative comparison with the baselines in the Yumi's Cells~\cite{yumicells} and Danbooru~\cite{danbooru2017} validation set. The scores for the CDR are reported only in the local regions along the edges, within the kernel size of 7.} 
\label{tab:sketch_q}
\end{table*}

\section{Implementation Details}
\label{sec:implementation_details}
This section provides the training details for edge enhancement, detailed architecture of edge-enhancing network, and hyper-parameters used in $S_{\text{pseudo}}$ generation, complementary to Section \ref{implementation_details}.
Afterward, hyper-parameters of the super-pixel methods for our proposed CDR are explained as well. Additionally, the implementation details of the sketch colorization are explained, complementary to the Section \ref{sec:05-sketch_colorization}.\vspace{0.2cm}

\noindent \textbf{Training Details for Edge-Enhancing Network}
We train and evaluate our model with a fixed size of 256 $\times$ 256 images for every dataset. 
We apply three edge-enhancing networks on the $5^{th}, 10^{th}$ and $17^{th}$ encoder layers of Zhang \etal~\cite{zhang2017real}.
In Su \etal~\cite{su2020insta}, instance-level colorization branch takes the patches of images cropped by their bounding boxes after the object detection module.
These object-level colorized outputs are fused into a full-image colorization branch via fusion modules, predicting the final colors.
For the full-image branch, we provide the pseudo-scribbles for the full image, as we do in Zhang \etal.
To accommodate our interactions into this object-level colorization branch as well, we crop the pseudo-scribbles as well as the images by their bounding boxes and give them to the branch as inputs.
Therefore, our edge-enhancing networks applied in this branch take the cropped scribbles corresponding to the cropped patches of color-bleeding artifacts, generating the refined activations to be fused into the full-image branch.
Both three edge-enhancing networks are applied on the $5^{th}, 10^{th}$ and $17^{th}$ encoder layers of instance-level and full-image network, respectively.
In the training phase, we set the hyper-parameters for each loss function as $\lambda_{\text{edge}} = \lambda_{\text{con}}$ $=$ 50 and $\lambda_{\text{reg}_1} = \lambda_{\text{reg}_2} = \lambda_{\text{reg}_3} = 1$ in Zhang \etal, and $\lambda_{\text{edge}}$ $=$ 50 and $\lambda_{\text{con}}$ $=$ $\lambda_{\text{reg}_1} = \lambda_{\text{reg}_2} = \lambda_{\text{reg}_3}$ $=$ 10 in Su \etal.
The width of the augmentation module for the $S_{pseudo}$ is randomly sampled from 1 to 10 pixels in the training phase.
We use Adam~\cite{kingma2014adam} optimizer with $\beta_1=0.9$ and $\beta_2=0.999$.
The learning rate is initially set to 0.01 and is gradually decayed for every epoch.

\noindent \textbf{Edge-Enhancing Network Architecture}
Edge-enhancing network consists of 4 convolutional layers, each of which contains a 3$\times$3 convolution filter with a stride of 1, ReLU~\cite{nair2010relu} and Batch normalization layer~\cite{ioffe2015batchnorm}.

\noindent \textbf{Pseudo-Scribble Generation} 
For generating the plausible approximation of real user-provided scribbles, we tune the hyper-parameters of the Canny edge extractor~\cite{canny1986cannyedge}\footnote[1]{Canny edge-extractor consists of 5 steps, which include noise reduction, Sobel filtering, non-maximum suppression, double threshold and edge tracking.\vspace{-0.5cm}} on every dataset, including ImageNet~\cite{imagenet_cvpr09}, COCO-Stuff~\cite{caesar2018coco}, Place205~\cite{zhou2017places}, Yumi's Cells~\cite{yumicells} and Danbooru~\cite{danbooru2017}, written in Table~\ref{tab:hyper_params_for_canny}.
We report Sigma ($\sigma$), high-threshold (TH$_h$), low-threshold (TH$_l$), and threshold gaps (TH$_{gap}$) of each dataset.
Sigma stands for the standard deviation of the Gaussian kernel for the noise reduction step.
Both TH$_h$ and TH$_l$ denote the threshold values for the double threshold step after the non-maximum suppression.
We apply different high thresholds for $I_{\text{gt},ab}$ and $I_{\text{init},ab}$ to select highly probable edges for the bleeding artifacts. 
In other words, we apply a rigid criterion on the ground-truth image for extracting the edges compared to the generated outputs, resulting in severely weak edges from this comparison.
We denote this threshold gap as TH$_{gap}$.

\noindent \textbf{Cluster Discrepancy Ratio}
We use simple linear iterative clustering~\cite{achanta2012slic} (SLIC) method to assign each pixel with a cluster assignment based on its colors and textures.
To focus on the color information, we run the SLIC on $ab$ channels of both ground-truth and colorized outputs.
We set the number of clusters to $250$, compactness to $10$, and sigma to $1$ for each image.
After we compute each ratio from the $a$ and $b$ channels and average them to produce the final score.

\noindent \textbf{Sketch Colorization}
We adjust the part of architecture and the training details of Zhang \etal~\cite{zhang2017real} to enable its colorization with local hints in the sketch colorization task.
We replace $1$) the input image from gray-scale to sketch image, and $2$) output channel size from 2 for $ab$ channels to 3 for RGB outputs.
To obtain the sketch image from the color image of each dataset, we first apply Gaussian blurring ($\sigma=0.7)$ to remove noisy edges and utilize a widely used edge extractor algorithm called XDoG~\cite{winnemoller2012xdog}, as used in Lee \etal~\cite{lee2020refsketchC} for the sketch colorization task.
Afterward, we obtain 54,317 training images and 6,036 test images for Yumi's Cells~\cite{yumicells}, and 7,014 and 380 images for Danbooru~\cite{danbooru2017}.
Then, we apply the training details proposed in the original paper~\cite{zhang2017real}, such as providing color hints and objective functions.
Note that the objective functions for color prediction are adjusted from minimizing the difference of $ab$ channels to RGB channels between generated output and the ground-truth. 
To train our model on this network, we convert the both generated RGB outputs images and ground-truth into \textit{Lab} images to apply our proposed objective edge-enhancing loss.

\begin{table}
\begin{center}
\begin{tabular}{@{}lclllclll@{}}
\toprule
Datasets & $\sigma$ & TH$_h$ & TH$_l$ & TH$_{gap}$\\
\midrule
ImageNet~\cite{imagenet_cvpr09} & 1.2 & \multicolumn{1}{c}{0.7} & \multicolumn{1}{c}{0.2} & \multicolumn{1}{c}{0.4}\\
COCO-Stuff~\cite{caesar2018coco} & 1.2 & \multicolumn{1}{c}{0.7} & \multicolumn{1}{c}{0.2} & \multicolumn{1}{c}{0.4}\\
Place205~\cite{zhou2017places} & 1.2 & \multicolumn{1}{c}{0.7} & \multicolumn{1}{c}{0.2} & \multicolumn{1}{c}{0.4}\\
Yumi's Cells~\cite{yumicells} & 1.3 & \multicolumn{1}{c}{0.7} & \multicolumn{1}{c}{0.2} & \multicolumn{1}{c}{0.4}\\
Danbooru~\cite{danbooru2017} & 0.7 & \multicolumn{1}{c}{0.8} & \multicolumn{1}{c}{0.2} & \multicolumn{1}{c}{0.5}\\
\bottomrule
\end{tabular}
\end{center}
\caption{Hyper-parameters for the Canny edge-extractor.}
\label{tab:hyper_params_for_canny}
\end{table}

\end{document}